\documentclass{article} 
\usepackage{iclr2025_conference,times}

\usepackage{wrapfig}
\usepackage{lipsum} 
\usepackage{wrapfig}

\usepackage{amsmath,amsfonts,bm}
\usepackage{multicol}

\def\eqref#1{equation~\ref{#1}}

\def\1{\bm{1}}

\DeclareMathAlphabet{\mathsfit}{\encodingdefault}{\sfdefault}{m}{sl}
\SetMathAlphabet{\mathsfit}{bold}{\encodingdefault}{\sfdefault}{bx}{n}

\usepackage{booktabs}      
\usepackage{multirow}      
\usepackage{amssymb}       

\usepackage{colortbl}
\usepackage{xcolor}

\usepackage{hyperref}
\usepackage{url}
\usepackage{multicol}
\usepackage{aliascnt}
\usepackage{adjustbox}
\usepackage{siunitx} 
\usepackage{booktabs}
\usepackage{multirow} 
\usepackage{enumitem}
\usepackage{booktabs} 
\usepackage{amssymb}  
\usepackage{amsmath}  
\usepackage{graphicx} 
\graphicspath{{./figs/}}
\usepackage{longtable} 
\usepackage{subcaption}
\usepackage{calc}
\usepackage[most]{tcolorbox}
\usepackage{listings}

\definecolor{uclablue}{rgb}{0.15, 0.45, 0.68}
\hypersetup{
    breaklinks,
    citecolor=uclablue,
    colorlinks=true,
}

\newtcolorbox{AIbox}[2][]{aibox,title=#2,#1}
\tcbset{
  aibox/.style={
    top=10pt,
    colframe=black,
    colbacktitle=black,
    coltitle=white,
    center,
  }
}

\lstdefinelanguage{prompt}{
    basicstyle=\scriptsize\ttfamily, 
    mathescape=true,        
    escapebegin=\color{latentcolor},  
    escapeend={},
    escapechar=@,
    stringstyle = \color{myorange},
    showstringspaces = false,
    moredelim = [s][\color{mypink}]{`}{`},
    moredelim = [s][\color{mybrown}]{```json}{```},
    moredelim = [s][\color{latentcolor}]{<StartOfLatent>}{<EndOfLatent>},
    literate = %
        {\ \ a.\ }{{\textcolor{mypurple}{\ \ a.\ }}}5
        {\ \ b.\ }{{\textcolor{mypurple}{\ \ b.\ }}}5
        {\ \ c.\ }{{\textcolor{mypurple}{\ \ c.\ }}}5
        {\ \ d.\ }{{\textcolor{mypurple}{\ \ d.\ }}}5
        {\ \ e.\ }{{\textcolor{mypurple}{\ \ e.\ }}}5
        {\ \ f.\ }{{\textcolor{mypurple}{\ \ f.\ }}}5
        {\ \ g.\ }{{\textcolor{mypurple}{\ \ g.\ }}}5
        {\ \ h.\ }{{\textcolor{mypurple}{\ \ h.\ }}}5
        {\ I.\ }{{\textcolor{mypurple}{\ I.\ }}}4
        {\ II.\ }{{\textcolor{mypurple}{\ II.\ }}}5
        {\ III.\ }{{\textcolor{mypurple}{\ III.\ }}}6
        {\ IV.\ }{{\textcolor{mypurple}{\ IV.\ }}}5
        {\ V.\ }{{\textcolor{mypurple}{\ V.\ }}}4
}

\newtcblisting[auto counter, number within=subsection]{prompt}[4][]{%
  before skip=6pt, after skip=6pt,    
  top=6pt, bottom=6pt, left=7pt, right=7pt, 
  enhanced, breakable,
  colback=white, colframe=gray!65,
  boxrule=0.6pt, arc=2pt,
  drop shadow={black!15},             
  listing only,
  listing options={
    language=prompt,
    upquote=true,
    basicstyle=\scriptsize\ttfamily \setlength{\baselineskip}{1.1\baselineskip},
    columns=fullflexible,
    breaklines=true,
    breakindent=0pt,
    xleftmargin=\ifx\empty#2\empty0pt\else#2\fi,
    xrightmargin=\ifx\empty#3\empty0pt\else#3\fi,
    aboveskip=0pt,                    
    belowskip=0pt,                    
    showstringspaces=false,
  },
  title={\ifstrempty{#4}{Prompt \thetcbcounter}{Prompt \thetcbcounter: #4}},
  colbacktitle=gray!6, coltitle=black,
  attach boxed title to top left={yshift=-1mm, xshift=6pt},
  boxed title style={colback=gray!6, boxrule=0pt, sharp corners},
  #1
}

\newtcblisting[auto counter, number within=subsection]{showcase}[4][]{%
  before=\par\vspace{\baselineskip},
  after=\par,
  width=\linewidth,
  enhanced,
  arc=0em,
  boxrule=1pt,
  listing only,
  listing options={
    language=prompt,
    upquote=true,
    basicstyle=\scriptsize\ttfamily \setlength{\baselineskip}{1.1\baselineskip},
    breaklines=true,
    breakindent=0pt,
    xleftmargin=\ifx\empty#2\empty-12pt\else#2\fi,
    xrightmargin=\ifx\empty#3\empty-5pt\else#3\fi,
    aboveskip=-4pt,
    belowskip=-4pt,
    columns=fullflexible,
    },
  colback=white,
  colframe=gray,
  colbacktitle=gray!5,
  coltitle=black,
  attach boxed title to top center={yshift=-3mm},
  box align=center,
  parbox=false,
  title={\ifstrempty{#4}{Example \thetcbcounter}{Example \thetcbcounter: #4}},
  #1
}

\definecolor{linkColor}{rgb}{0.2,0.4,0.6}
\definecolor{myblue}{HTML}{0379AC}
\definecolor{myred}{HTML}{A50E50}
\definecolor{myorange}{RGB}{238, 133, 74}
\definecolor{latentcolor}{named}{cyan}
\definecolor{normalcolor}{RGB}{0, 0, 0}
\usepackage{marvosym}

\usepackage{tabularx}
\usepackage{bbm}
\usepackage{makecell}

\usepackage{cleveref}
\newcommand{\minisection}[1]{\noindent{\textbf{#1}}.}

\usepackage{caption}
\title{Scaling Native Multimodal Pre-Training From Scratch}

\author{
Haoyuan Wu$^{1, 2}$,
Aoqi Wu$^{2}$,
Hai Wang$^{2}$,
Jiajia Wu$^{2}$,
Jinxiang Ou$^{2}$,
Bei Yu$^{1}$
\\
\textbf{$^1$The Chinese University of Hong Kong} \quad 
\textbf{$^2$LLM Department, Tencent}
}

\usepackage{needspace}
\definecolor{RevisionBlue}{HTML}{0758A8}
\definecolor{OriginalGray}{HTML}{555555}
\definecolor{ReviewOrange}{HTML}{9A4E00}
\newif\ifshoworiginal
\showoriginalfalse 

\begin{document}
\maketitle

\begin{abstract}

Although large language models (LLMs) exhibit remarkable reasoning capabilities, their reliance on text-only pre-training restricts the perception of the multimodal physical world.
Native multimodal pre-training avoids this limitation by training models from scratch on multimodal inputs, thereby achieving deep cross-modal integration and mitigating optimization asymmetries inherent to traditional late-fusion architectures.
Despite these advantages, the scaling properties of this paradigm remain incompletely characterized. 
To address this gap, we investigate the optimal model size and token count for training a Transformer-based vision-language model under a fixed computational budget. 
Our study demonstrates that minimal objective loss adheres to a predictable compute law, whereas compute-optimal model sizes and token counts scale as power laws.
Notably, language and multimodal objectives manifest distinct allocation trends.
The language allocation exponents lie in a similar range across the different data mixtures.
The multimodal model-allocation exponent decreases modestly with the multimodal token ratio, indicating a relative shift toward token allocation.
Additionally, our scaling analysis yields a budget-compensation rule. 
Specifically, an additional multimodal-token budget can offset the text-efficiency penalty caused by incorporating visual information into native multimodal pre-training under a fixed compute budget.
Downstream evaluations further reveal that native multimodal pre-training is associated with improved spatial reasoning and multimodal few-shot learning.
Generally, this empirical research establishes the essential groundwork for predictably scaling multimodal foundation models. 

\end{abstract}

\section{Introduction}

Large language models (LLMs)~\citep{deepmind2026gemini,anthropic2026claude,openai2026gpt} demonstrate remarkable reasoning and generation capabilities, establishing themselves as powerful foundation models. 
Despite these successes, text-only pre-training remains constrained by its inability to ground multimodal concepts in the physical world.
Multimodal pre-training addresses this limitation by incorporating data from multiple modalities.

Currently, the dominant paradigm for multimodal pre-training relies on late fusion~\citep{shukor2025scalingnmm}. 
Typically, this approach couples a pre-trained language model with a vision encoder~\citep{radford2021clip,zhai2023siglip,tschannen2025siglip2} via a projection layer, followed by continued training on multimodal data~\citep{kimi2026kimi25,liu2023llava,lin2026moellava}.
Although this strategy efficiently reuses pre-trained weights, it introduces a fundamental asymmetry.
Specifically, vision and language representations are learned independently from distinct data distributions and under different optimization objectives.

To resolve this asymmetry, researchers explore native multimodal pre-training~\citep{shukor2025scalingnmm,cui2025emu35}.
In this paradigm, models are trained from scratch on multimodal data, allowing vision and language representations to develop jointly.
This approach raises a practical question: how should a fixed compute budget be allocated between model size and training tokens? 
For unimodal language models, this resource allocation is guided by compute-optimal scaling laws that dictate how to distribute a budget between model size and data~\citep{kaplan2020oaiscaling,hoffmann2022chinchilla}.
However, it remains unclear whether native multimodal pre-training follows analogous scaling laws and whether the optimal allocations for language and multimodal objectives align or conflict.

To bridge this gap, we establish compute-optimal scaling laws for native multimodal pre-training.
Specifically, we derive these laws using two estimators, IsoFLOP profiles and training-curve envelopes~\citep{hoffmann2022chinchilla}.
These estimators provide complementary views of the compute-optimal allocation.
We demonstrate that the minimum loss for each objective follows a power law with a fitted offset, while compute-optimal model sizes and token counts scale as power laws~\citep{kaplan2020oaiscaling,hoffmann2022chinchilla}.

Fitting scaling laws separately for the language and multimodal objectives reveals different allocation trends across different data mixtures.
The language allocation exponents remain in a similar range.
For the multimodal objective, the model-allocation exponent decreases modestly as the multimodal token ratio increases, indicating relatively greater token allocation. 

Furthermore, our central question is whether the extra multimodal compute budget can compensate for the text-efficiency penalty incurred by incorporating visual information under a fixed compute budget. 
We estimate this requirement by inverting jointly fitted text-loss compute frontiers with a common offset and decay exponent.
The resulting compensation rule expresses the required total compute as a mixture-dependent multiple of the pure-text reference budget.
Across different data mixtures, the estimated compute increase is smaller than the increase associated with retaining the reference model size and text-token count and adding multimodal tokens.

Beyond scaling dynamics, we also evaluate the downstream implications of native multimodal pre-training across language and multimodal understanding tasks.
Our findings indicate that native multimodal pre-training improves pure-text spatial reasoning, demonstrating that spatial understanding acquired through multimodal training successfully generalizes to unimodal text tasks.
Moreover, this paradigm exhibits multimodal in-context learning capabilities analogous to those observed in traditional LLMs~\citep{brown2020gpt3}.

These results provide empirical guidance for scaling native multimodal pre-training.
Our main contributions are summarized as follows:
\begin{itemize}[leftmargin=2em]
\item We independently analyze the scaling behaviors of language and multimodal objectives, revealing objective-specific allocation trends across different data mixtures.
\item We derive a budget-compensation rule showing that the estimated additional compute needed to match pure-text loss is smaller than the compute increase associated with adding multimodal tokens.
\item We empirically demonstrate that native multimodal pre-training improves pure-text spatial reasoning and enables robust multimodal in-context learning. 
\end{itemize}
\section{Preliminaries}

\subsection{Compute-Optimal Allocation}

In this work, we revisit a fundamental problem in native multimodal pre-training: given a fixed computational budget $C$, how should resources be optimally allocated between the model size $N$ (the number of activated non-embedding parameters) and the number of training tokens $D$?
Furthermore, we investigate how this optimal trade-off is influenced by data composition, parameterized by the multimodal data ratio $r$.
To formalize this problem, we model the final pre-training loss $L(N, D)$ as a function of $N$ and $D$.
Using the standard approximation $C = 6ND$, the computational budget is strictly determined by these two variables.
Our objective is to minimize the loss under a fixed compute constraint:
\begin{equation}
    N^{\mathrm{opt}}(C), D^{\mathrm{opt}}(C) = \underset{N,D \ \mathrm{s.t.}\ \mathrm{FLOPs}(N,D)=C}{\arg\min} L(N,D),
\label{eq:alloc}
\end{equation}
where $N^{\mathrm{opt}}(C)$ and $D^{\mathrm{opt}}(C)$ denote the compute-optimal allocation for the given budget $C$. 
Given the lack of solid multimodal validation metrics, and because each token is seen roughly once during pre-training, we rely solely on the smoothed training loss as a proxy for the test loss, ensuring standardized evaluation across both language and multimodal objectives.
In native multimodal pre-training, the language and multimodal objectives ($L_{\mathrm{text}}$ and $L_{\mathrm{mm}}$) are optimized simultaneously using shared model parameters. 
It remains unclear whether the two objectives follow similar scaling laws or favor different compute allocations.
To investigate this dynamic, we estimate compute-optimal allocation separately for each objective using the same jointly trained models.
With $r=D_{\mathrm{mm}}/D_{\mathrm{text}}$, define $C_s=6ND_s$ for $s\in\{\mathrm{text},\mathrm{mm}\}$. 
Thus $D_{\mathrm{total}}=(1+r)D_{\mathrm{text}}$ and $C_{\mathrm{total}}=C_{\mathrm{text}}+C_{\mathrm{mm}}=(1+r)C_{\mathrm{text}}$. 
These are objective-specific compute coordinates; at $r=0$, the total compute equals $C_{\mathrm{text}}$.
For both objectives, the compute-optimal model size and token count follow power laws:
\begin{equation}
N_s^{\mathrm{opt}}(C_s;r)\propto C_s^{a_s(r)},
\qquad
D_s^{\mathrm{opt}}(C_s;r)\propto C_s^{b_s(r)},
\qquad
a_s(r)+b_s(r)=1.
\label{eq:alloc_modality}
\end{equation}

\subsection{Estimators of the Compute-Optimal Frontier}

We estimate compute-optimal allocation for each objective in~\Cref{eq:alloc} using two methods.
The IsoFLOP profile method serves as our primary estimator, while the training-curve envelope provides a sensitivity check on the same training runs. 
For budget compensation, we invert the $L_{\mathrm{text}}$ frontier to compare the compute required to match pure-text loss with the budget associated with adding multimodal tokens.
Budget selection is specified in~\Cref{appendix:preprocessing}, and the estimators are detailed in~\Cref{appendix:allocation_estimators}.

\minisection{IsoFLOP profiles}
At a predetermined compute budget $C$, plotting the loss of each model against $\log N$ generates an IsoFLOP profile. 
A quadratic fit in $\log N$ estimates the preferred model size $N^{\mathrm{opt}}(C)$, subject to the observed size range.
The optimal token count then follows algebraically as $D^{\mathrm{opt}}(C)=C/(6N^{\mathrm{opt}})$. 
Regressing $\log N^{\mathrm{opt}}$ and $\log D^{\mathrm{opt}}$ on $\log C$ yields the allocation exponents $a$ and $b$.
The fitted minimum interpolates across the evaluated model sizes at each budget.

\minisection{Training curve envelope}
For a fixed model size $N$, increasing $D$ during training produces a trajectory of loss versus compute $C=6ND$. 
Pooling the trajectories of all evaluated models and extracting the lower envelope yields the lowest observed loss at each evaluated budget $C$. 
Every point on this envelope corresponds to a specific $(N, D)$ configuration.
Therefore, regressing $\log N$ and $\log D$ against $\log C$ across these points generates a second estimate of $a$ and $b$ from the same runs, providing a sensitivity check on the IsoFLOP fits.
We provide more details in \Cref{appendix:scaling}.

\minisection{The compute frontier $L(C)$}
Along the lower envelope, we fit loss with a power law in compute:
\begin{equation}
L_s(C_s;r)
= E_s(r)
+ \left(\frac{C_{c,s}(r)}{C_s}\right)^{\beta_s(r)},
\qquad s\in\{\mathrm{text},\mathrm{mm}\},
\label{eq:power_law}
\end{equation}
where $E_s(r)$ is a fitted loss offset, $C_{c,s}(r)$ is a characteristic compute scale, and $\beta_s(r)$ is a decay exponent distinct from the token-allocation exponent $b_s$.
We fit these parameters using the procedure in~\Cref{appendix:scaling}. 
This power law describes the frontier as a function of compute within the observed range.

\subsection{Experimental Setup}

\minisection{Training data}
Our model is trained on a mixture of text and multimodal datasets.
The corpus contains 250B text tokens derived from web pages, books, academic papers, and other domains. 
Moreover, we construct a large-scale multimodal dataset comprising web-crawled image-text pairs and interleaved image-text documents.
With images represented as continuous patch embeddings, the multimodal corpus contains 75B tokens.

\minisection{Model architecture}
We employ an MoE architecture based on a decoder-only Transformer. 
A single patch embedding layer maps images directly to continuous embeddings without a pre-trained vision encoder. 
Notably, we train these MoE models using an auxiliary-loss-free approach~\citep{liu2024deepseekv3}.
More implementation details are provided in~\Cref{appendix:imple}.
\begin{figure}[!tb]
\centering
\includegraphics[width=0.995\linewidth]{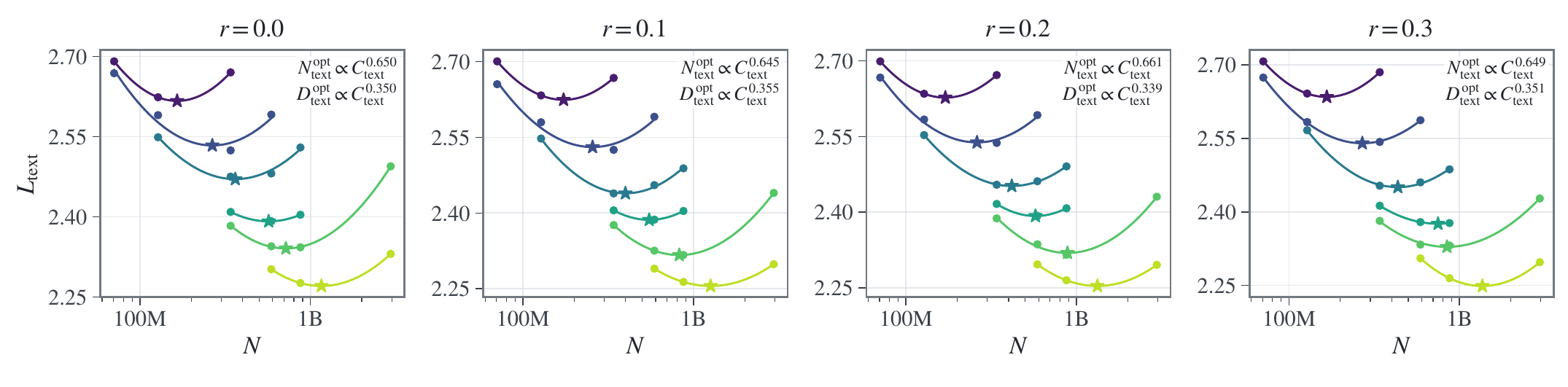}
\caption{\textbf{IsoFLOP curves (language objective).}
At each target budget $C_{\mathrm{text}}$, we read the $L_{\mathrm{text}}$ of every model size at the token count satisfying $6ND_{\mathrm{text}}=C_{\mathrm{text}}$ and plot it against $\log N$.
The fitted minima estimate compute-optimal model sizes within the observed range and determine the corresponding token counts.
}
\label{fig:text_isoflop}
\end{figure}

\begin{figure}[!tb]
\centering
\includegraphics[width=0.995\linewidth]{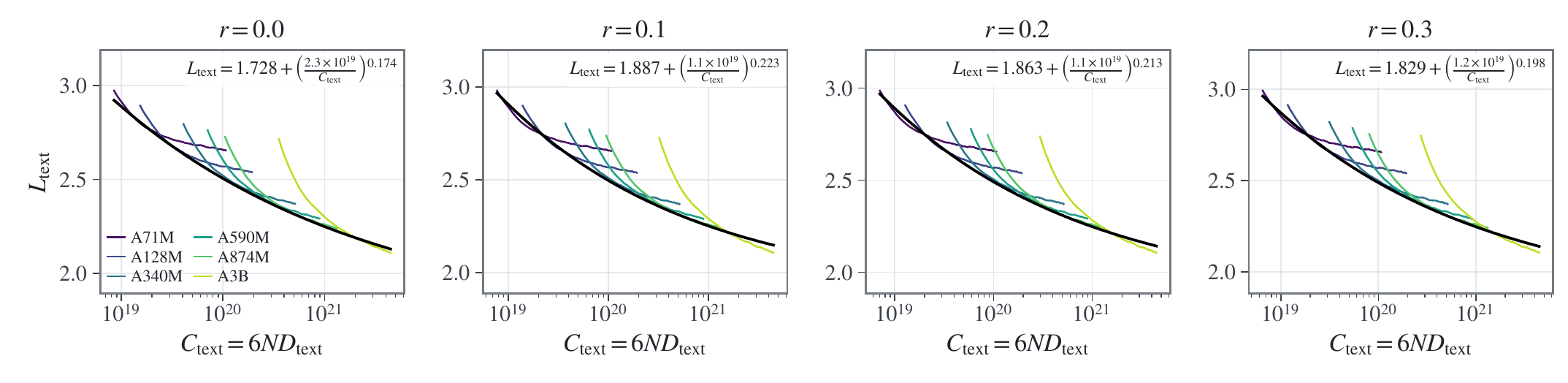}
\caption{\textbf{Training curve envelope (language objective).} 
Training curves span the evaluated model sizes and mixture ratios $r$.
For each ratio, the lower envelope gives the lowest observed loss at each budget $C_{\mathrm{text}}$; its selected model sizes and token counts determine the allocation estimates. The fitted loss frontier summarizes this envelope.
}
\label{fig:text_envelop}
\end{figure}

\section{Scaling Native Multimodal Pre-Training}
\label{sec:scaling}
Building on our established estimators, we investigate the compute-optimal scaling behavior of native multimodal models. 
By decoupling the resource allocation problem into isolated objectives, we first examine how the mixture ratio $r$ affects compute-optimal allocation for each objective. 
We then compare the additional compute needed to match pure-text loss with the compute increase associated with adding multimodal tokens.

\subsection{The Language Objective}

\begin{figure}[!tb]
\centering
\includegraphics[width=0.995\linewidth]{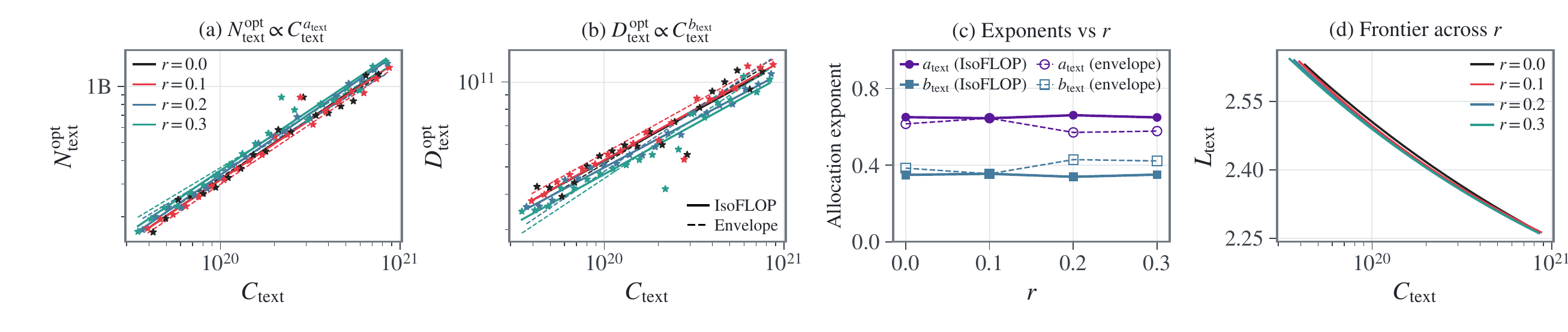}
\caption{\textbf{Compute-optimal allocation (language objective).} (a) Optimal model size $N_{\mathrm{text}}^{\mathrm{opt}}$ and (b) optimal token count $D_{\mathrm{text}}^{\mathrm{opt}}$ as a function of the compute budget $C_{\mathrm{text}}$ for each $r$. 
Solid lines and stars represent the IsoFLOP fits and their respective minima, while thin dashed lines and scattered points provide a cross-check using the training envelope. 
(c) Allocation exponents $a_{\mathrm{text}}(r)$ and $b_{\mathrm{text}}(r)$ from both estimators. 
(d) Fitted compute frontiers $L_{\mathrm{text}}(C_{\mathrm{text}};r)$.
The two estimators give model-allocation exponents in a similar range across the measured ratios.
}
\label{fig:text_alloc}
\end{figure}

\begin{figure}[!tb]
\centering
\includegraphics[width=0.765\linewidth]{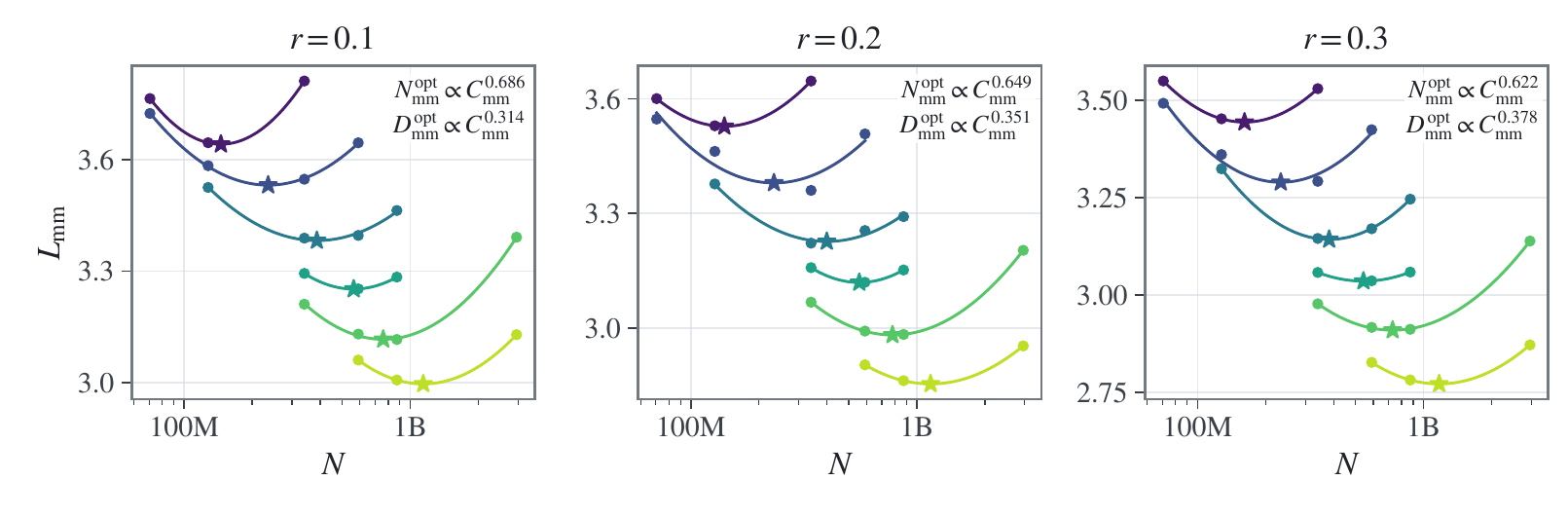}
\caption{\textbf{IsoFLOP curves (multimodal objective).}
At each target budget $C_{\mathrm{mm}}$, we read the $L_{\mathrm{mm}}$ of every model size at the token count satisfying $6ND_{\mathrm{mm}}=C_{\mathrm{mm}}$ and plot it against $\log N$.
The fitted minima estimate compute-optimal model sizes within the observed range and determine the corresponding token counts.
}
\label{fig:mm_isoflop}
\end{figure}

We first isolate the language objective $L_{\mathrm{text}}$ under the text-compute constraint $C_{\mathrm{text}}$. 
As shown in~\Cref{fig:text_isoflop,fig:text_envelop}, the IsoFLOP profiles and training-curve envelopes reveal predictable loss frontiers across mixtures. 
The IsoFLOP allocation exponents remain close to $a_{\mathrm{text}}\simeq0.65$ and $b_{\mathrm{text}}\simeq0.35$, indicating a similar balance of model capacity and text tokens across different $r$. 
\Cref{fig:text_alloc} compares the two estimators; the fitting procedures are described in~\Cref{appendix:scaling}.

\subsection{The Multimodal Objective}

Applying the same estimators to $L_{\mathrm{mm}}$ under $C_{\mathrm{mm}}$ reveals a shift in allocation. 
From $r=0.1$ to $0.3$, the IsoFLOP model-allocation exponent decreases from 0.686 to 0.622, while the token-allocation exponent increases from 0.314 to 0.378.
The envelope estimates show the same direction of change.
These results indicate a modest shift toward token allocation as the multimodal-to-text token ratio increases within the evaluated range.

\subsection{Budget Compensation at a Fixed $L_{\mathrm{text}}$ Target}
\label{sec:budget_compensation}

We now quantify the additional compute required for native multimodal pre-training to match a pure-text loss target.
We compare this requirement with the compute budget associated with adding multimodal tokens while retaining the reference model size and text-token count.

\begin{figure}[!tb]
\centering
\includegraphics[width=0.765\linewidth]{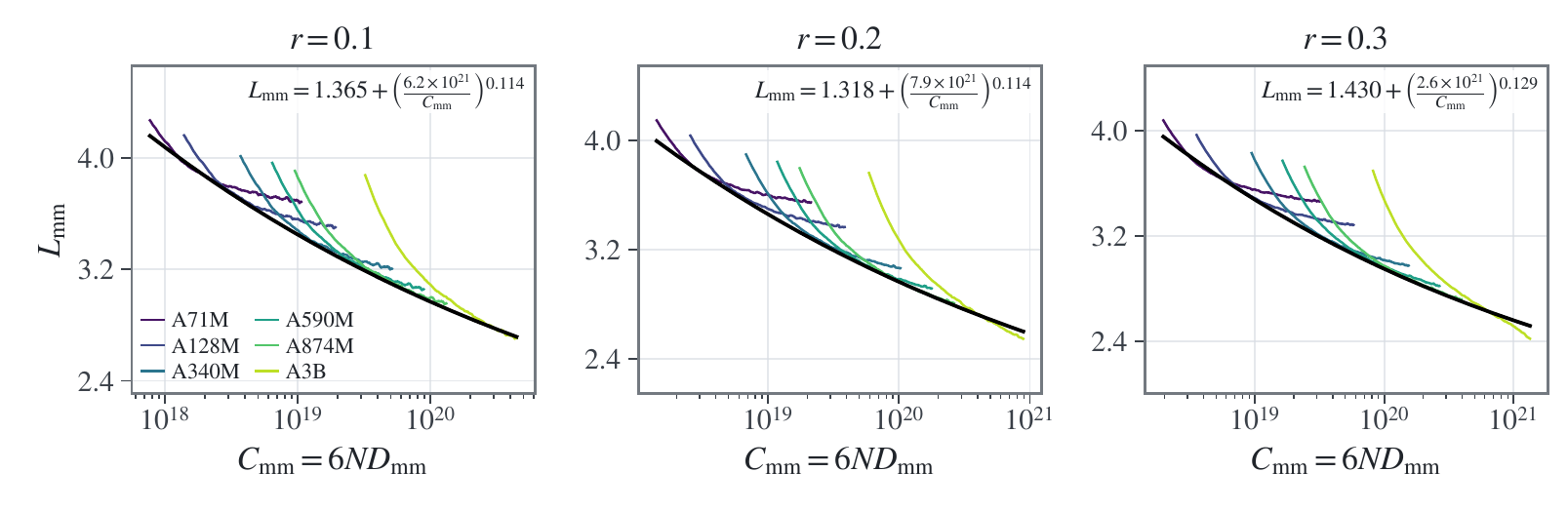}
\caption{\textbf{Training curve envelope (multimodal objective).} 
Training curves span the evaluated model sizes and mixture ratios $r$.
For each ratio, the lower envelope gives the lowest observed loss at each budget $C_{\mathrm{mm}}$; its selected model sizes and token counts determine the allocation estimates. The fitted loss frontier summarizes this envelope.
}
\label{fig:mm_envelop}
\end{figure}

\begin{figure}[!tb]
\centering
\includegraphics[width=0.995\linewidth]{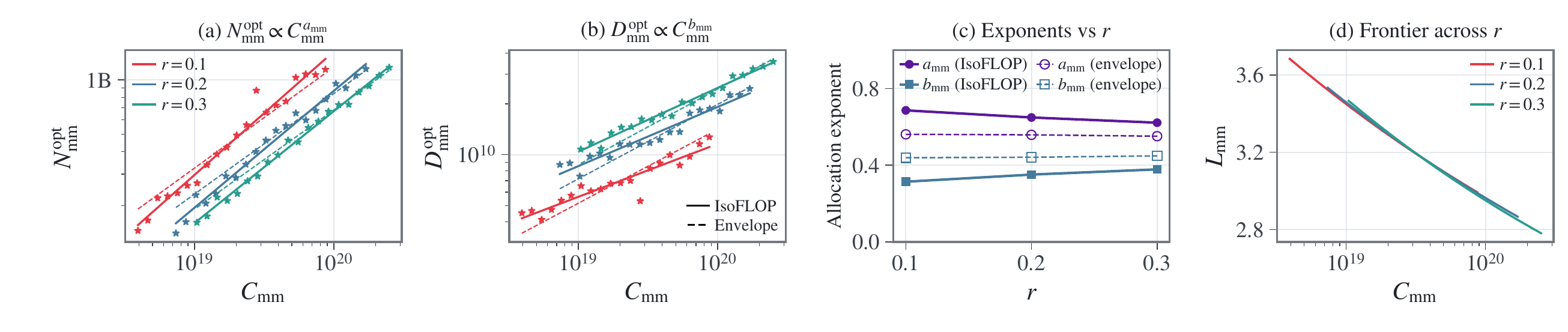}
\caption{\textbf{Compute-optimal allocation (multimodal objective).} (a) Optimal model size $N_{\mathrm{mm}}^{\mathrm{opt}}$ and (b) optimal token count $D_{\mathrm{mm}}^{\mathrm{opt}}$ as a function of the compute budget $C_{\mathrm{mm}}$ for each $r$. 
Solid lines and stars represent the IsoFLOP fits and their respective minima, while thin dashed lines and scattered points provide a cross-check using the training envelope. 
(c) Allocation exponents $a_{\mathrm{mm}}(r)$ and $b_{\mathrm{mm}}(r)$ from both estimators. (d) Fitted compute frontiers $L_{\mathrm{mm}}(C_{\mathrm{mm}};r)$. The model-allocation exponents decrease modestly across different $r$.
}
\label{fig:mm_alloc}
\end{figure}

\begin{figure}[!tb]
\centering
\includegraphics[width=0.975\linewidth]{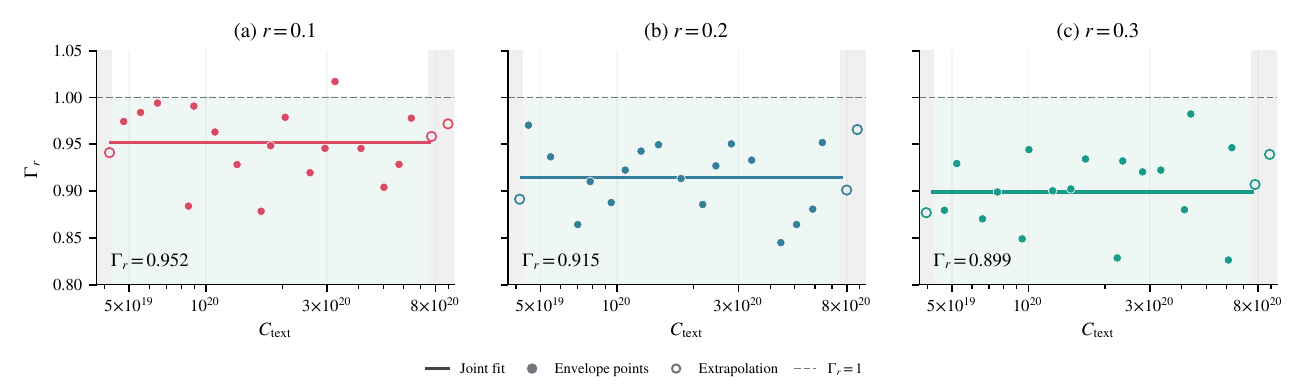}
\caption{\textbf{Budget compensation under the joint frontier fit.} 
Lines show the ratio $\Gamma_r$ of required to available total compute; values below one indicate that the available budget exceeds the fitted requirement.
Points transform empirical envelope losses using the fitted pure-text reference; open points and gray regions indicate reference extrapolation. 
The fitted factors remain below one over the common supported range.
}
\label{fig:budget_envelope_points}
\end{figure}

\minisection{A scaling rule for compensation}
To obtain a compensation rule, we jointly fit the text-loss frontiers with a common offset and decay exponent and a separate compute scale for each mixture:
\begin{equation}
\overline L_{\mathrm{text}}(C_{\mathrm{text}};r)=\overline E_{\mathrm{text}}+
\left(\frac{\overline C_{c,\mathrm{text}}(r)}{C_{\mathrm{text}}}\right)^{\overline\beta_{\mathrm{text}}}.
\label{eq:shared_shape_frontier}
\end{equation}
Bars distinguish this joint fit from the separate fits in~\Cref{eq:power_law}. 
The offset $\overline E_{\mathrm{text}}$ and exponent $\overline\beta_{\mathrm{text}}$ are constant across mixtures, so differences between frontiers are represented by $\overline C_{c,\mathrm{text}}(r)$.
The preceding figures use the separate frontier fits; the joint fitting procedure is described in~\Cref{appendix:compensation_models}.

For the pure-text reference ($r=0$), write $C_0=C_{\mathrm{text}}$ and set the target loss to $L_{\mathrm{text}}^\star=\overline L_{\mathrm{text}}(C_0;0)$.
Retaining the reference model size and text-token count and adding $rD_{\mathrm{text}}$ multimodal tokens increases the available total compute to $(1+r)C_0$ under the approximation $C=6ND$.
The mixture's text-compute requirement, obtained by inverting the joint frontier, is
\begin{equation}
\widehat C_{\mathrm{text}}(L_{\mathrm{text}}^\star;r)
=\overline C_{c,\mathrm{text}}(r)
\bigl(L_{\mathrm{text}}^\star-\overline E_{\mathrm{text}}\bigr)^{-1/\overline\beta_{\mathrm{text}}}.
\end{equation}
Thus, $\widehat C_{\mathrm{total}}=(1+r)\widehat C_{\mathrm{text}}$, and the ratio of required to available total compute is
\begin{equation}
\Gamma_r=\frac{\widehat C_{\mathrm{total}}(L_{\mathrm{text}}^\star;r)}{(1+r)C_0}
=\frac{\overline C_{c,\mathrm{text}}(r)}{\overline C_{c,\mathrm{text}}(0)}.
\label{eq:added_token_coverage}
\end{equation}
The added compute is sufficient when $\Gamma_r\leq1$, or equivalently when
\begin{equation}
	\widehat C_{\mathrm{total}}(L_{\mathrm{text}}^\star;r)-C_0\leq rC_0.
\end{equation}
Considering that the offset and decay exponent are constant across different data mixtures, $\Gamma_r$ is independent of the target loss under this model. 

\minisection{Compensation across mixtures}
As shown in~\Cref{fig:budget_envelope_points}, we obtain $\Gamma_r\approx0.952$, $0.915$, and $0.899$ at $r=0.1,0.2,0.3$, respectively. 
The corresponding total-compute requirements, $(1+r)\Gamma_r C_0$, exceed the pure-text budget by approximately 4.7\%, 9.8\%, and 16.9\%. 
These increases are smaller than the available compute increases of 10\%, 20\%, and 30\%, respectively. 
Thus, for the fitted values,
\begin{equation}
	0<\widehat C_{\mathrm{total}}(L_{\mathrm{text}}^\star;r)-C_0<rC_0.
\end{equation}

The fitted frontiers therefore imply a text-efficiency cost in total compute, which is smaller than the budget increase associated with adding multimodal tokens. 
Under the joint model, the absolute additional compute required to match the target is $[(1+r)\Gamma_r-1]C_0$.

\minisection{Observed variation across mixtures}
Across the different $r$, $\Gamma_r$ decreases from 0.952 to 0.899, indicating a larger relative margin within the available budget. 
However, the total-compute requirement relative to pure-text training, $(1+r)\Gamma_r$, increases from 1.047 to 1.169. 
A smaller $\Gamma_r$ therefore does not imply a lower total cost. 

This rule compares compute requirements along the model-family frontiers within their common fitted range. 
The available budget is defined by retaining the reference model size and text-token count, but achieving the fitted minimum may require a different allocation of that budget. 
The result therefore does not guarantee loss recovery at a fixed model size or under a fixed text-token constraint. 
Points outside the pure-text reference range in~\Cref{fig:budget_envelope_points} require extrapolation.

\section{Downstream Implications}
\label{sec:exp}

\begin{figure}[!tb]
\centering
\includegraphics[width=0.60\linewidth]{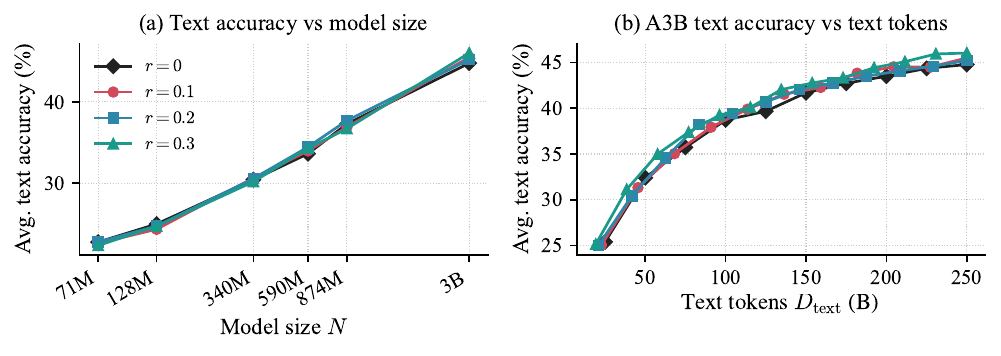}
\caption{\textbf{Text capabilities are preserved under native multimodal pre-training.} Average accuracy across 16 text benchmarks with varying multimodal data ratios $r$ (fixed 250B text-token budget). (a) Text performance across model sizes $N$. (b) A3B text performance over training tokens $D_{\mathrm{text}}$. The curves indicate broadly comparable aggregate performance at matched text-token budgets.
}
\label{fig:text_under_mm}
\end{figure}

\subsection{Evaluation of Base Models}

\minisection{Text capabilities}
The evaluation of text capabilities is conducted using 16 distinct benchmarks.
\begin{itemize}[itemsep=0pt, topsep=0pt, parsep=0pt, leftmargin=2em]
    \item \textbf{Aggregate:} SuperGPQA\citep{du2025supergpqa}(5-shot), MMLU-Redux\citep{hendrycks2020mmlu}(5-shot), MMLU-Pro\citep{wang2024mmlupro}(5-shot), and AGIEval$_{\mathrm{en}}$\citep{zhong2023agieval}(3-shot).
    \item \textbf{Coding:} HumanEval+\citep{liu2023evalplus}(0-shot) and MBPP+\citep{liu2023evalplus}(3-shot).
    \item \textbf{Mathematics:} GSM8K\citep{cobbe2021gsm8k}(4-shot, CoT) and MATH\citep{lightman2023math500}(4-shot, CoT).
    \item \textbf{Logical Reasoning:} BBH\citep{suzgun2022bbh}(3-shot, CoT) and SpatialEval\citep{wang2024spatialeval}(1-shot).
    \item \textbf{Knowledge:} NaturalQuestions\citep{kwiatkowski2019naturalquestions}~(5-shot) and TriviaQA\citep{joshi2017triviaqa}(5-shot).
    \item \textbf{Commonsense Reasoning:} SIQA\citep{sap2019socialiqa}(0-shot), PIQA\citep{bisk2020piqa}(0-shot), HellaSwag\citep{zellers2019hellaswag}(10-shot), and WinoGrande\citep{sakaguchi2021winogrande}(5-shot).
\end{itemize}

\minisection{Multimodal capabilities}
The evaluation of multimodal capabilities is conducted using 23 distinct benchmarks.
\begin{itemize}[itemsep=0pt, topsep=0pt, parsep=0pt, leftmargin=2em]
    \item \textbf{Aggregate:} MMStar\citep{chen2024mmstar}(3-shot), MMMU\citep{yue2024mmmu}(3-shot), MMMU-Pro\citep{yue2025mmmupro}(3-shot), MME\citep{fu2026mme}(3-shot), and MMBench$_{\mathrm{en}}$\citep{liu2024mmbench}(3-shot).
    \item \textbf{VQA:} VQAv2\citep{goyal2017vqav2}(1-shot) and TextVQA\citep{singh2019textvqa}(1-shot).
    \item \textbf{STEM:} MathVista\citep{lu2024mathvista}(3-shot), MathVerse\citep{zhang2024mathverse}(3-shot), and ScienceQA\citep{lu2022sciencqa}(3-shot, CoT).
    \item \textbf{Doc Understanding:} HallusionBench\citep{guan2024hallusionbench}(3-shot), LogicVista\citep{xiao2024logicvista}(1-shot, CoT), AI2D\citep{kembhavi2016ai2d}(3-shot), and ChartQA\citep{masry2022chartqa}(3-shot).
    \item \textbf{Vision Knowledge:} MMBench$_{\mathrm{cc}}$\citep{liu2024mmbench}(3-shot) and SimpleVQA\citep{cheng2025simplevqa}(3-shot).
    \item \textbf{Counting:} CountBench(0-shot)\citep{paiss2023countbench} and CountQA\citep{tamarapalli2025countqa}(1-shot).
    \item \textbf{Spatial Reasoning:} RealWorldQA\citep{grok2025realworldqa}(3-shot), CV-Bench\citep{tong2024cvbench}(3-shot), OmniSpatial\citep{jia2025omnispatial}(3-shot), SEAM\citep{tang2025seam}(3-shot), and SpatialEval\citep{wang2024spatialeval}(1-shot).
\end{itemize}

\minisection{Evaluation protocol}
For native multimodal pre-training, we evaluate the pre-trained models in an in-context learning setting. 
For multiple-choice tasks, we report accuracy by selecting the option that yields the lowest perplexity. 
Meanwhile, open-ended tasks are evaluated using the exact-match metric against reference answers, whereas coding tasks are evaluated using the Pass@1 metric.

\minisection{Image processing}
To accommodate the 4K-token context limit of our pre-trained models, we restrict the number of visual tokens per image. 
Specifically, images are partitioned into $32 \times 32$ patches and proportionally downscaled if the resulting grid exceeds the token budget. 
We allocate a maximum of 1536 tokens per image, reducing this limit to 512 tokens in few-shot scenarios to ensure both the templates and the test query fit within the context window.

\minisection{Few-Shot examples}
To ensure representative coverage, few-shot templates are sampled across distinct categories from each benchmark's development or training split. 
These templates are formatted as interleaved image-question-answer sequences and prepended to the test query.

\begin{figure}[!tb]
\centering
\includegraphics[width=0.60\linewidth]{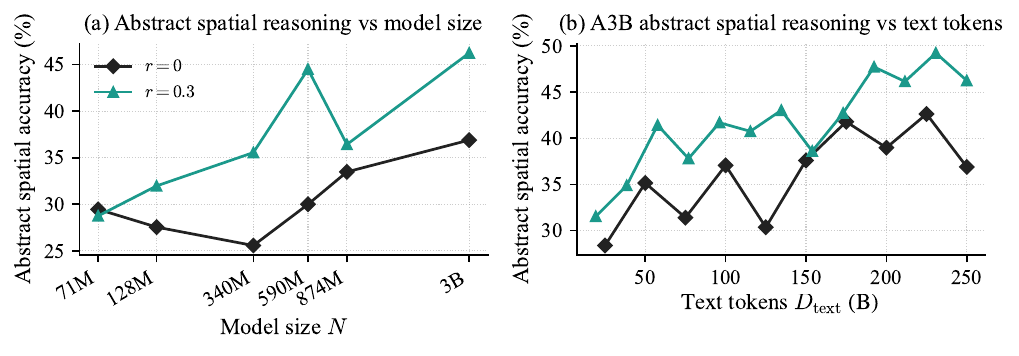}
\caption{\textbf{Multimodal pre-training enhances pure-text spatial reasoning.} Accuracy on SpatialEval's text-only abstract spatial-reasoning subtasks~\citep{wang2024spatialeval}, comparing baseline ($r=0$) and multimodal ($r=0.3$) settings. (a) Accuracy across model sizes $N$. (b) A3B accuracy over training tokens $D_{\mathrm{text}}$. Multimodal models consistently outperform text-only baselines, with the performance gap widening at larger scales.}
\label{fig:spatial_abstract_under_mm}
\end{figure}

\subsection{Text Performance under Multimodal Pre-Training}
\label{sec:text_under_mm}

A central concern in native multimodal pre-training is whether it compromises core language capabilities. 
To compare models at matched text exposure, we fix the text budget at 250B tokens and vary the multimodal budget from 0B (text-only, $r=0$) to 75B ($r=0.3$), evaluating models across six scales ($N$ ranging from 71M to 3B).

\minisection{Preservation of core language capabilities}
Introducing multimodal data leaves aggregate text performance broadly comparable at matched text-token budgets.
As shown in~\Cref{fig:text_under_mm}, average accuracy across 16 text benchmarks remains consistent across all multimodal data ratios. 
This parity holds across the entire model family.
Furthermore, models trained with different data ratios exhibit nearly identical performance trajectories throughout A3B training. 
These aggregate results show no large text-performance deterioration under this protocol.

\minisection{Cross-modal transfer in spatial reasoning}
The most compelling evidence of cross-modal transfer emerges in the abstract spatial reasoning subtasks of SpatialEval~\citep{wang2024spatialeval} (MazeNav and SpatialMap). 
Although these queries are strictly text-based and lack visual input, incorporating multimodal tokens during pre-training yields substantial performance gains.
As illustrated in~\Cref{fig:spatial_abstract_under_mm}, multimodal models ($r=0.3$) consistently outperform text-only baselines ($r=0$).
Notably, this advantage scales with model size. While smaller models (71M and 128M) exhibit marginal differences, the performance gap widens substantially for larger models up to 3B.
This suggests that spatial and relational structures learned from visual modalities are embedded within the shared parameter space, successfully generalizing to enhance unimodal text comprehension.

\subsection{Multimodal Few-Shot Learning}

Beyond aggregate accuracy, we also investigate whether native multimodal pre-training endows base models with multimodal in-context learning.
This allows models to leverage a few image-text templates during inference without parameter updates, analogous to text-only LLMs.
We evaluate this across our entire model family, with model size $N$ ranging from 71M to 3B. 
All models are trained on 250B text tokens and 75B multimodal tokens. 
Evaluations are conducted in 0-, 1-, and 3-shot settings. 
Given the instability of CoT reasoning in pre-trained base models under the zero-shot setting, we conduct our analysis of multimodal few-shot learning on 21 multimodal benchmarks.
Full results are detailed in~\Cref{table:fewshot_0shot,table:fewshot_1shot,table:fewshot_3shot}.

\begin{figure}[!tb]
\centering
\includegraphics[width=0.60\linewidth]{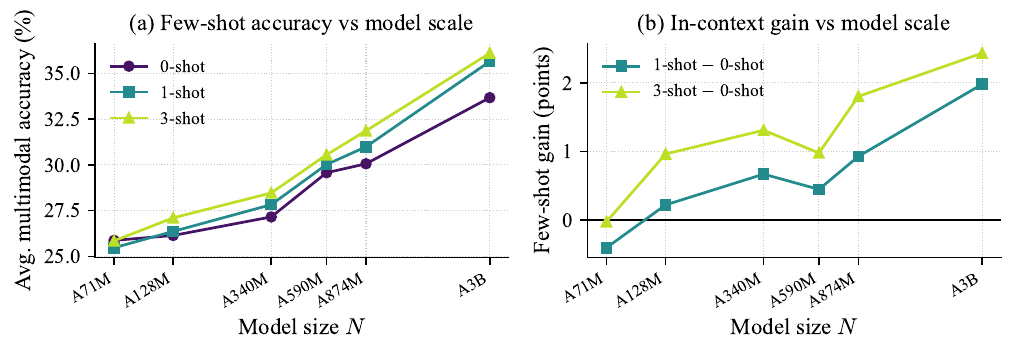}
\caption{\textbf{Multimodal in-context learning emerges with model scaling.}
(a) Average multimodal accuracy for 0-, 1-, and 3-shot settings across model sizes $N$. (b) Relative performance gains of $k$-shot settings over the 0-shot baseline. The increasing gains demonstrate that multimodal in-context learning emerges with model scaling.
}
\label{fig:fewshot_scale}
\end{figure}

\minisection{In-context learning emerges with model scaling}
\Cref{fig:fewshot_scale} demonstrates that the efficacy of in-context templates increases alongside model scale.
At the smallest scale, these templates provide no measurable benefit; the 3-shot average is virtually identical to the 0-shot baseline, and a 1-shot evaluation slightly degrades performance.
However, the performance gain scales positively with model capacity, increasing from near zero for the A71M model to +1.80 points for the A874M model, and ultimately reaching +2.43 points at the A3B scale.
Furthermore, at the largest scale, aggregated accuracy consistently improves with shots.
This progression mirrors the in-context learning capabilities observed in text-only LLMs, indicating that in-context learning capabilities emerge naturally from native multimodal pre-training.

\begin{figure}[!tb]
\centering
\includegraphics[width=0.70\linewidth]{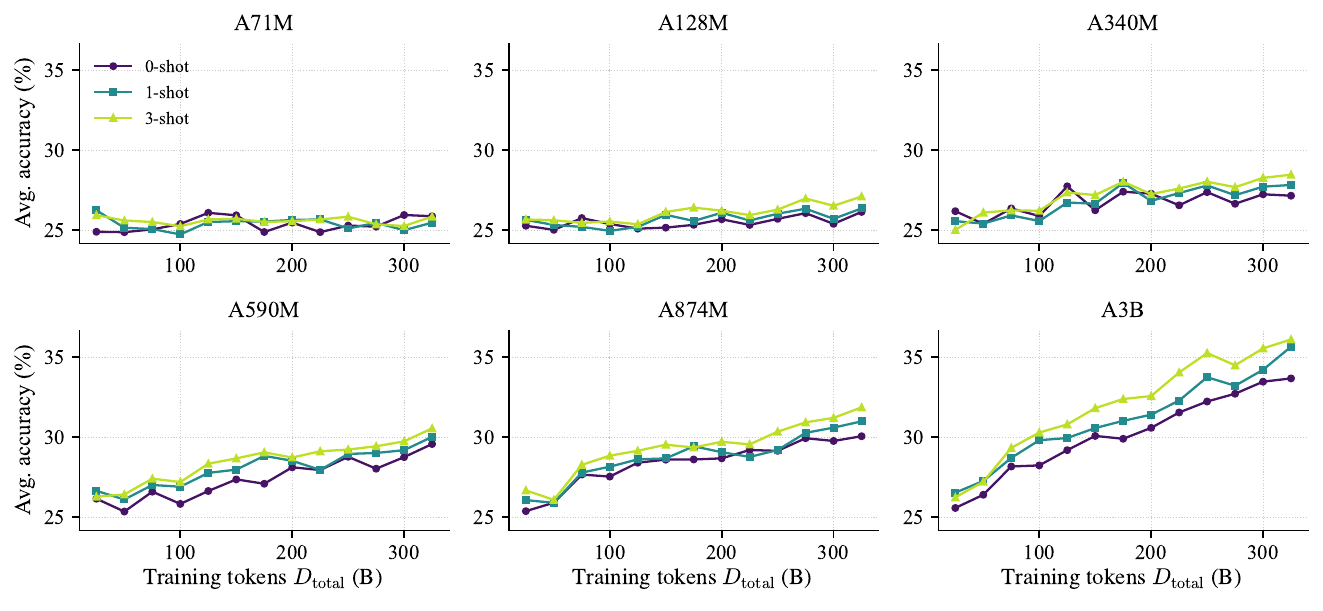}
\caption{\textbf{Few-shot accuracy under data scaling.}
Average benchmark accuracy of each model in 0-, 1-, and 3-shot settings across total training tokens $D_{\mathrm{total}}$. 
Although the performance trajectories remain overlapping throughout training for smaller models, the few-shot settings progressively outperform the 0-shot baseline in larger models.
}
\label{fig:fewshot_train}
\end{figure}

\minisection{In-context learning strengthens with data scaling}
\Cref{fig:fewshot_train} illustrates a similar trend across each model's training process. 
The performance margin afforded by few-shot prompting is initially absent; early in training, the 0-, 1-, and 3-shot evaluation curves are virtually indistinguishable. 
The expected hierarchical ordering, where an increased number of shots yields higher accuracy, only emerges during later stages of training.
The onset of this divergence depends heavily on model capacity. 
For the A71M through A340M models, the performance curves remain overlapping throughout the entire training process. 
Conversely, for the A874M and A3B models, a clear performance gap emerges early and widens steadily as the training progresses. 
This indicates that in-context learning is acquired gradually, requiring both sufficient model scale and substantial training data to manifest.

\begin{figure}[!tb]
\centering
\includegraphics[width=0.70\linewidth]{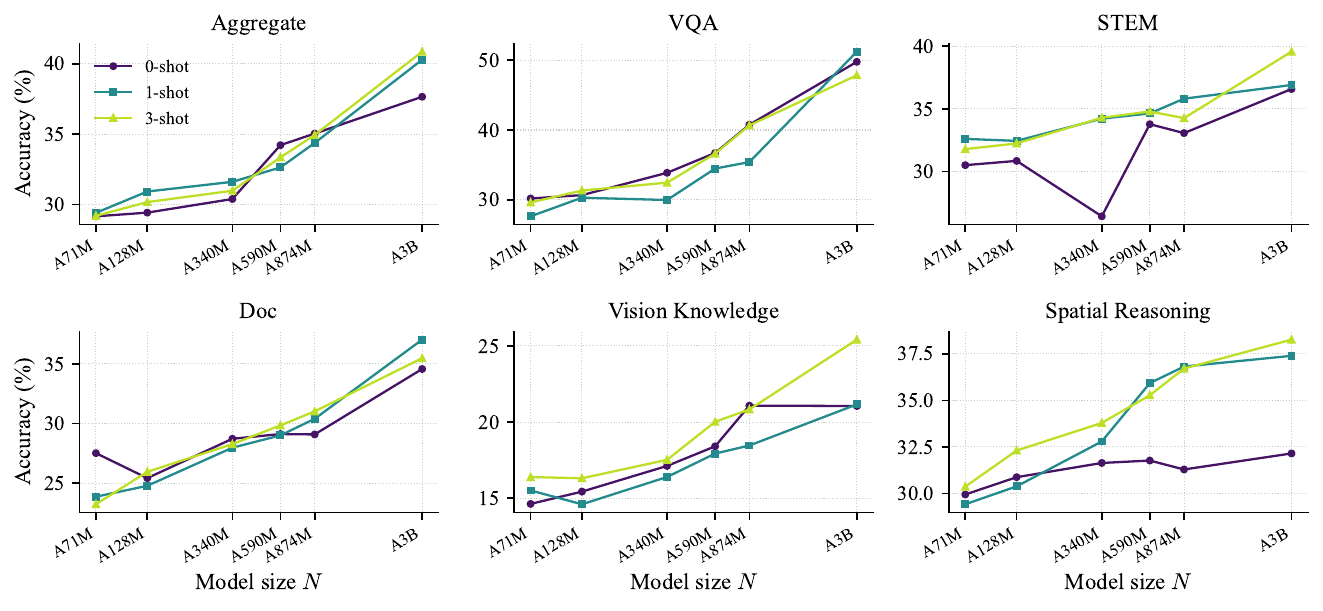}
\caption{\textbf{Few-shot trends by task category across model scale.}
Each panel aggregates the benchmarks of one category and reports 0-, 1-, and 3-shot accuracy across model size $N$.
The few-shot margin is largest and most consistent on spatial reasoning, modest on aggregate and vision-knowledge suites, and absent or negative on OCR/recognition benchmarks.
}
\label{fig:fewshot_category}
\end{figure}

\minisection{Few-shot gains concentrate on spatial reasoning}
\Cref{fig:fewshot_category} demonstrates that the benefits of in-context learning vary considerably across task categories.
The most pronounced and consistent improvements are observed in spatial-reasoning benchmarks, alongside notable gains in structured-diagram and aggregate evaluation suites.
Conversely, performance on OCR- and recognition-oriented benchmarks plateaus or even degrades as in-context templates are introduced.
This suggests that the templates primarily assist the model in resolving the spatial and relational structure of a query, rather than enhancing fine-grained visual recognition or text extraction.
This phenomenon aligns with the cross-modal spatial-reasoning transfer discussed in~\Cref{sec:text_under_mm}. 
A detailed per-benchmark breakdown is provided in~\Cref{appendix:exp}.

\begin{figure}[!tb]
\centering
\includegraphics[width=0.60\linewidth]{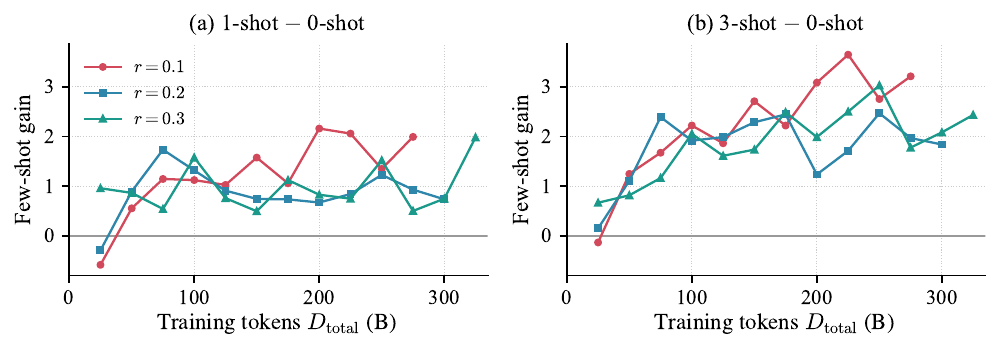}
\caption{\textbf{Few-shot gain trends upward with training.}
Few-shot gain of the A3B model over the 0-shot baseline for the 1-shot and 3-shot settings, shown for $r\in\{0.1,0.2,0.3\}$ against total training tokens $D_{\mathrm{total}}$.
Despite checkpoint-to-checkpoint fluctuations, the gains show an overall positive trend with $D_{\mathrm{total}}$ across all $r$.
}
\label{fig:fewshot_mmtoken}
\end{figure}

\minisection{Few-shot gain trends upward with training}
\Cref{fig:fewshot_mmtoken} plots the A3B few-shot gain against total training tokens $D_{\mathrm{total}}$ for each $r$. 
Although the per-checkpoint gains fluctuate, they show an overall upward trend as training progresses, and this holds for all three values of $r$. 
The 3-shot gains are generally larger than the 1-shot gains. 
Since the three $r$ curves trend upward together, we read the few-shot gain as positively correlated with $D_{\mathrm{total}}$ rather than tied to any particular $r$.
Together with \Cref{fig:fewshot_train}, this suggests that multimodal in-context learning strengthens as the model sees more training data.

\section{Related Work}

\minisection{Multimodal pre-training paradigms}
Most vision-language models adopt a late-fusion paradigm, coupling a pre-trained language model with a separately pre-trained vision encoder via a lightweight projection layer for subsequent adaptation on multimodal data~\citep{radford2021clip,zhai2023siglip,tschannen2025siglip2,liu2023llava,lin2026moellava,kimi2026kimi25}.
Although this modular design efficiently reuses robust unimodal checkpoints, it introduces a fundamental asymmetry.
Vision and language representations are learned independently from disparate data distributions and distinct training objectives, thereby constraining deep cross-modal integration.
To address this limitation, emerging research investigates native multimodal pre-training, where a single model is trained from scratch on interleaved multimodal data to establish a unified representational space from the outset~\citep{shukor2025scalingnmm,cui2025emu35,tong2026beyondlang}.
Our work adopts this paradigm by eliminating the vision encoder; instead, a single projection layer maps images into continuous patch embeddings for direct processing by a decoder-only Transformer.
We focus on compute-optimal scaling of native multimodal models.

\minisection{Compute-optimal scaling laws}
Scaling laws characterize the predictable relationship between model performance, scale, training data, and compute, offering principled guidance for resource allocation under fixed training budgets.
For unimodal language models, \citet{kaplan2020oaiscaling} first establish power-law relationships between loss and scale. 
Subsequently, \citet{hoffmann2022chinchilla} refine the compute-optimal frontier using IsoFLOP profiles and training-curve envelopes, demonstrating that model size and training tokens should scale proportionally.
These frameworks are the standard methodology for performance extrapolation and compute allocation.
However, existing analyses predominantly focus on unimodal settings with fixed data distributions. 
Consequently, compute-optimal allocation for jointly optimizing heterogeneous modalities within one model remains incompletely characterized.
\citet{shukor2025scalingnmm} investigate scaling behaviors for native multimodal models, and they model the training objective as a single aggregate loss.
In contrast, we decouple the language and multimodal objectives and employ both IsoFLOP profiles and training-curve envelopes as complementary estimation methods~\citep{hoffmann2022chinchilla,shukor2026datamixscaling}.
Our findings describe objective-specific allocation trends across the different data mixtures.
Inverting the text-loss frontiers also quantifies the compute required to match a pure-text loss target at the evaluated ratios.
\section{Conclusion}

In this work, we establish compute-optimal scaling laws for native multimodal pre-training, systematically characterizing its scaling behavior. 
By estimating separate frontiers for the language and multimodal objectives, we demonstrate predictable scaling with different allocation trends across data mixtures.
Language allocation exponents remain in a similar range, while the multimodal objective favors relatively greater token allocation as the multimodal token ratio increases.
Under the joint frontier model, the additional compute required to match pure-text loss is smaller than the compute increase associated with retaining the reference model size and text-token count and adding multimodal tokens. 
Beyond scaling, our experiments show that native multimodal pre-training retains broadly comparable aggregate text performance at matched text-token budgets, with spatial-reasoning improvements and multimodal few-shot learning.
These findings connect compute allocation, text-loss budget compensation, and downstream performance to guide native multimodal pre-training.
\clearpage
{
\bibliographystyle{iclr2025_conference}
\bibliography{ref/Top, ref/reference}

@string{aaai     = "Proc.~AAAI"}

@string{acl      = "Proc.~ACL"}

@string{colm     = "Proc.~COLM"}

@string{iclr     = "Proc.~ICLR"}

@string{icml     = "Proc.~ICML"}

@string{nips     = "Proc.~NIPS"}

@string{cvpr     = "Proc.~CVPR"}

@string{eccv     = "Proc.~ECCV"}

@string{iccv     = "Proc.~ICCV"}

@string{open     = "Optimization and Engineering"}

@string{arxiv    = "arXiv preprint"}

@string{science   = "Science"}

@string{aaai     = "AAAI Conference on Artificial Intelligence (AAAI)"}

@string{acl      = "Annual Meeting of the Association for Computational Linguistics (ACL)"}

@string{colm     = "Conference on Language Modeling (COLM)"}

@string{iclr     = "International Conference on Learning Representations (ICLR)"}

@string{icml     = "International Conference on Machine Learning (ICML)"}

@string{nips     = "Annual Conference on Neural Information Processing Systems (NIPS)"}

@string{cvpr     = "IEEE Conference on Computer Vision and Pattern Recognition (CVPR)"}

@string{eccv     = "European Conference on Computer Vision (ECCV)"}

@string{iccv     = "IEEE International Conference on Computer Vision (ICCV)"}

@article{liu2024deepseekv3,
  title={{DeepSeek-V3 Technical Report}},
  author={Liu, Aixin and Feng, Bei and Xue, Bing and Wang, Bingxuan and Wu, Bochao and Lu, Chengda and Zhao, Chenggang and Deng, Chengqi and Zhang, Chenyu and Ruan, Chong and others},
  journal={arXiv preprint arXiv:2412.19437},
  year={2024}
}

@article{kimi2026kimi25,
  title={{Kimi K2.5: Visual Agentic Intelligence}},
  author={Kimi and Bai, Tongtong and Bai, Yifan and Bao, Yiping and Cai, SH and Cao, Yuan and Charles, Y and Che, HS and Chen, Cheng and Chen, Guanduo and others},
  journal={arXiv preprint arXiv:2602.02276},
  year={2026}
}

@misc{anthropic2026claude,
  author = {{Anthropic}},
  title = {{Claude Opus 4.8}},
  howpublished = {\url{https://www.anthropic.com/claude/opus}},
  year = {2026},
}

@misc{deepmind2026gemini,
  author = {{Google DeepMind}},
  title = {{Gemini 3.1 Pro}},
  howpublished = {\url{https://deepmind.google/technologies/gemini/pro}},
  year = {2026},
}

@misc{openai2026gpt,
  author = {{OpenAI}},
  title = {{GPT 5.5}},
  howpublished = {\url{https://openai.com/index/introducing-gpt-5-5}},
  year = {2026},
}

@inproceedings{radford2021clip,
  title={{Learning Transferable Visual Models From Natural Language Supervision}},
  author={Radford, Alec and Kim, Jong Wook and Hallacy, Chris and Ramesh, Aditya and Goh, Gabriel and Agarwal, Sandhini and Sastry, Girish and Askell, Amanda and Mishkin, Pamela and Clark, Jack and others},
  booktitle=icml,
  year={2021},
}

@inproceedings{zhai2023siglip,
  title={{Sigmoid Loss for Language Image Pre-Training}},
  author={Zhai, Xiaohua and Mustafa, Basil and Kolesnikov, Alexander and Beyer, Lucas},
  booktitle=iccv,
  year={2023}
}

@article{tschannen2025siglip2,
  title={{SigLIP 2: Multilingual Vision-Language Encoders with Improved Semantic Understanding, Localization, and Dense Features}},
  author={Tschannen, Michael and Gritsenko, Alexey and Wang, Xiao and Naeem, Muhammad Ferjad and Alabdulmohsin, Ibrahim and Parthasarathy, Nikhil and Evans, Talfan and Beyer, Lucas and Xia, Ye and Mustafa, Basil and others},
  journal={arXiv preprint arXiv:2502.14786},
  year={2025}
}

@article{hendrycks2020mmlu,
  title={{Measuring Massive Multitask Language Understanding}},
  author={Hendrycks, Dan and Burns, Collin and Basart, Steven and Zou, Andy and Mazeika, Mantas and Song, Dawn and Steinhardt, Jacob},
  journal={arXiv preprint arXiv:2009.03300},
  year={2020}
}

@inproceedings{wang2024mmlupro,
  title={{MMLU-Pro: A More Robust and Challenging Multi-Task Language Understanding Benchmark}},
  author={Wang, Yubo and Ma, Xueguang and Zhang, Ge and Ni, Yuansheng and Chandra, Abhranil and Guo, Shiguang and Ren, Weiming and Arulraj, Aaran and He, Xuan and Jiang, Ziyan and others},
  booktitle=nips,
  year={2024}
}

@article{suzgun2022bbh,
  title={{Challenging Big-Bench Tasks and Whether Chain-of-Thought Can Solve Them}},
  author={Suzgun, Mirac and Scales, Nathan and Sch{\"a}rli, Nathanael and Gehrmann, Sebastian and Tay, Yi and Chung, Hyung Won and Chowdhery, Aakanksha and Le, Quoc V and Chi, Ed H and Zhou, Denny and others},
  journal={arXiv preprint arXiv:2210.09261},
  year={2022}
}

@article{du2025supergpqa,
  title={{SuperGPQA: Scaling LLM Evaluation Across 285 Graduate Disciplines}},
  author={Du, Xinrun and Yao, Yifan and Ma, Kaijing and Wang, Bingli and Zheng, Tianyu and Zhu, King and Liu, Minghao and Liang, Yiming and Jin, Xiaolong and Wei, Zhenlin and others},
  journal={arXiv preprint arXiv:2502.14739},
  year={2025}
}

@article{cobbe2021gsm8k,
  title={{Training Verifiers to Solve Math Word Problems}},
  author={Cobbe, Karl and Kosaraju, Vineet and Bavarian, Mohammad and Chen, Mark and Jun, Heewoo and Kaiser, Lukasz and Plappert, Matthias and Tworek, Jerry and Hilton, Jacob and Nakano, Reiichiro and others},
  journal={arXiv preprint arXiv:2110.14168},
  year={2021}
}

@inproceedings{lightman2023math500,
  title={{Let's Verify Step by Step}},
  author={Lightman, Hunter and Kosaraju, Vineet and Burda, Yuri and Edwards, Harrison and Baker, Bowen and Lee, Teddy and Leike, Jan and Schulman, John and Sutskever, Ilya and Cobbe, Karl},
  booktitle=iclr,
  year={2023}
}

@inproceedings{liu2023evalplus,
  title={{Is Your Code Generated by ChatGPT Really Correct? Rigorous Evaluation of Large Language Models for Code Generation}},
  author={Liu, Jiawei and Xia, Chunqiu Steven and Wang, Yuyao and Zhang, Lingming},
  booktitle=nips,
  volume={36},
  pages={21558--21572},
  year={2023}
}

@article{zhong2023agieval,
  title={{AGIEval: A Human-Centric Benchmark for Evaluating Foundation Models}},
  author={Zhong, Wanjun and Cui, Ruixiang and Guo, Yiduo and Liang, Yaobo and Lu, Shuai and Wang, Yanlin and Saied, Amin and Chen, Weizhu and Duan, Nan},
  journal={arXiv preprint arXiv:2304.06364},
  year={2023}
}

@article{kwiatkowski2019naturalquestions,
  title={{Natural Questions: A Benchmark for Question Answering Research}},
  author={Kwiatkowski, Tom and Palomaki, Jennimaria and Redfield, Olivia and Collins, Michael and Parikh, Ankur and Alberti, Chris and Epstein, Danielle and Polosukhin, Illia and Devlin, Jacob and Lee, Kenton and others},
  journal={Transactions of the Association for Computational Linguistics},
  volume={7},
  pages={453--466},
  year={2019},
  publisher={MIT Press One Rogers Street, Cambridge, MA 02142-1209, USA journals-info~…}
}

@article{joshi2017triviaqa,
  title={{TriviaQA: A Large Scale Distantly Supervised Challenge Dataset for Reading Comprehension}},
  author={Joshi, Mandar and Choi, Eunsol and Weld, Daniel S and Zettlemoyer, Luke},
  journal={arXiv preprint arXiv:1705.03551},
  year={2017}
}

@article{zellers2019hellaswag,
  title={{Hellaswag: Can A Machine Really Finish Your Sentence?}},
  author={Zellers, Rowan and Holtzman, Ari and Bisk, Yonatan and Farhadi, Ali and Choi, Yejin},
  journal={arXiv preprint arXiv:1905.07830},
  year={2019}
}

@inproceedings{bisk2020piqa,
  title={{PIQA: Reasoning about Physical Commonsense in Natural Language}},
  author={Bisk, Yonatan and Zellers, Rowan and Gao, Jianfeng and Choi, Yejin and others},
  booktitle=aaai,
  volume={34},
  number={05},
  pages={7432--7439},
  year={2020}
}

@article{sap2019socialiqa,
  title={{SocialiQA: Commonsense Reasoning about Social Interactions}},
  author={Sap, Maarten and Rashkin, Hannah and Chen, Derek and LeBras, Ronan and Choi, Yejin},
  journal={arXiv preprint arXiv:1904.09728},
  year={2019}
}

@article{sakaguchi2021winogrande,
  title={{Winogrande: An Adversarial Winograd Schema Challenge at Scale}},
  author={Sakaguchi, Keisuke and Bras, Ronan Le and Bhagavatula, Chandra and Choi, Yejin},
  journal={Communications of the ACM},
  volume={64},
  number={9},
  pages={99--106},
  year={2021},
  publisher={ACM New York, NY, USA}
}

@inproceedings{chen2024mmstar,
  title={{Are We on the Right Way for Evaluating Large Vision-Language Models?}},
  author={Chen, Lin and Li, Jinsong and Dong, Xiaoyi and Zhang, Pan and Zang, Yuhang and Chen, Zehui and Duan, Haodong and Wang, Jiaqi and Qiao, Yu and Lin, Dahua and others},
  booktitle=nips,
  year={2024}
}

@inproceedings{yue2024mmmu,
  title={{MMMU: A Massive Multi-Discipline Multimodal Understanding and Reasoning Benchmark for Expert AGI}},
  author={Yue, Xiang and Ni, Yuansheng and Zhang, Kai and Zheng, Tianyu and Liu, Ruoqi and Zhang, Ge and Stevens, Samuel and Jiang, Dongfu and Ren, Weiming and Sun, Yuxuan and others},
  booktitle=cvpr,
  year={2024}
}

@inproceedings{yue2025mmmupro,
  title={{MMMU-Pro: A More Robust Multi-Discipline Multimodal Understanding Benchmark}},
  author={Yue, Xiang and Zheng, Tianyu and Ni, Yuansheng and Wang, Yubo and Zhang, Kai and Tong, Shengbang and Sun, Yuxuan and Yu, Botao and Zhang, Ge and Sun, Huan and others},
  booktitle=acl,
  year={2025}
}

@inproceedings{fu2026mme,
  title={{MME: A Comprehensive Evaluation Benchmark for Multimodal Large Language Models}},
  author={Fu, Chaoyou and Chen, Peixian and Shen, Yunhang and Qin, Yulei and Zhang, Mengdan and Lin, Xu and Yang, Jinrui and Zheng, Xiawu and Li, Ke and Sun, Xing and others},
  booktitle=nips,
  year={2026}
}

@inproceedings{liu2024mmbench,
  title={{MMBench: Is Your Multi-Modal Model an All-Around Player?}},
  author={Liu, Yuan and Duan, Haodong and Zhang, Yuanhan and Li, Bo and Zhang, Songyang and Zhao, Wangbo and Yuan, Yike and Wang, Jiaqi and He, Conghui and Liu, Ziwei and others},
  booktitle=eccv,
  year={2024},
}

@inproceedings{lu2024mathvista,
  title={{MathVista: Evaluating Mathematical Reasoning of Foundation Models in Visual Contexts}},
  author={Lu, Pan and Bansal, Hritik and Xia, Tony and Liu, Jiacheng and Li, Chunyuan and Hajishirzi, Hannaneh and Cheng, Hao and Chang, Kai-Wei and Galley, Michel and Gao, Jianfeng},
  booktitle=iclr,
  year={2024}
}

@inproceedings{zhang2024mathverse,
  title={{MathVerse: Does Your Multi-Modal LLM Truly See the Diagrams in Visual Math Problems?}},
  author={Zhang, Renrui and Jiang, Dongzhi and Zhang, Yichi and Lin, Haokun and Guo, Ziyu and Qiu, Pengshuo and Zhou, Aojun and Lu, Pan and Chang, Kai-Wei and Qiao, Yu and others},
  booktitle=eccv,
  year={2024}
}

@inproceedings{lu2022sciencqa,
  title={{Learn to explain: Multimodal Reasoning via Thought Chains for Science Question Answering}},
  author={Lu, Pan and Mishra, Swaroop and Xia, Tanglin and Qiu, Liang and Chang, Kai-Wei and Zhu, Song-Chun and Tafjord, Oyvind and Clark, Peter and Kalyan, Ashwin},
  booktitle=nips,
  year={2022}
}

@inproceedings{guan2024hallusionbench,
  title={{Hallusionbench: An Advanced Diagnostic Suite for Entangled Language Hallucination and Visual Illusion in Large Vision-Language Models}},
  author={Guan, Tianrui and Liu, Fuxiao and Wu, Xiyang and Xian, Ruiqi and Li, Zongxia and Liu, Xiaoyu and Wang, Xijun and Chen, Lichang and Huang, Furong and Yacoob, Yaser and others},
  booktitle=cvpr,
  year={2024}
}

@article{xiao2024logicvista,
  title={{Logicvista: Multimodal LLM Logical Reasoning Benchmark in Visual Contexts}},
  author={Xiao, Yijia and Sun, Edward and Liu, Tianyu and Wang, Wei},
  journal={arXiv preprint arXiv:2407.04973},
  year={2024}
}

@inproceedings{kembhavi2016ai2d,
  title={{A Diagram Is Worth A Dozen Images}},
  author={Kembhavi, Aniruddha and Salvato, Mike and Kolve, Eric and Seo, Minjoon and Hajishirzi, Hannaneh and Farhadi, Ali},
  booktitle=eccv,
  year={2016}
}

@inproceedings{masry2022chartqa,
  title={{ChartQA: A Benchmark for Question Answering about Charts with Visual and Logical Reasoning}},
  author={Masry, Ahmed and Do, Xuan Long and Tan, Jia Qing and Joty, Shafiq and Hoque, Enamul},
  booktitle=acl,
  year={2022}
}

@inproceedings{goyal2017vqav2,
  title={{Making the V in VQA Matter: Elevating the Role of Image Understanding in Visual Question Answering}},
  author={Goyal, Yash and Khot, Tejas and Summers-Stay, Douglas and Batra, Dhruv and Parikh, Devi},
  booktitle=cvpr,
  year={2017}
}

@inproceedings{singh2019textvqa,
  title={{Towards VQA Models That Can Read}},
  author={Singh, Amanpreet and Natarajan, Vivek and Shah, Meet and Jiang, Yu and Chen, Xinlei and Batra, Dhruv and Parikh, Devi and Rohrbach, Marcus},
  booktitle=cvpr,
  year={2019}
}

@inproceedings{paiss2023countbench,
  title={{Teaching Clip to Count to Ten}},
  author={Paiss, Roni and Ephrat, Ariel and Tov, Omer and Zada, Shiran and Mosseri, Inbar and Irani, Michal and Dekel, Tali},
  booktitle=cvpr,
  year={2023}
}

@article{tamarapalli2025countqa,
  title={{CountQA: How Well Do MLLMs Count in the Wild?}},
  author={Tamarapalli, Jayant Sravan and Grover, Rynaa and Pande, Nilay and Yerramilli, Sahiti},
  journal={arXiv preprint arXiv:2508.06585},
  year={2025}
}

@misc{grok2025realworldqa,
  author = {{xAI}},
  title = {{RealWorldQA}},
  howpublished = {\url{https://huggingface.co/datasets/xai-org/RealworldQA}},
  year = {2024},
}

@inproceedings{tong2024cvbench,
  title={{Cambrian-1: A Fully Open, Vision-Centric Exploration of Multimodal LLMs}},
  author={Tong, Shengbang and Brown, Ellis and Wu, Penghao and Woo, Sanghyun and Middepogu, Manoj and Akula, Sai C and Yang, Jihan and Yang, Shusheng and Iyer, Adithya and Pan, Xichen and others},
  booktitle=nips,
  year={2024}
}

@article{jia2025omnispatial,
  title={{OmniSpatial: Towards Comprehensive Spatial Reasoning Benchmark for Vision Language Models}},
  author={Jia, Mengdi and Qi, Zekun and Zhang, Shaochen and Zhang, Wenyao and Yu, Xinqiang and He, Jiawei and Wang, He and Yi, Li},
  journal={arXiv preprint arXiv:2506.03135},
  year={2025}
}

@inproceedings{tang2025seam,
  title={{SEAM: Semantically Equivalent Across Modalities Benchmark for Vision-Language Models}},
  author={Tang, Zhenwei and Jiao, Difan and Yang, Blair and Anderson, Ashton},
  booktitle=colm,
  year={2025}
}

@inproceedings{wang2024spatialeval,
  title={{Is a Picture Worth a Thousand Words? Delving into Spatial Reasoning for Vision Language Models}},
  author={Wang, Jiayu and Ming, Yifei and Shi, Zhenmei and Vineet, Vibhav and Wang, Xin and Li, Yixuan and Joshi, Neel},
  booktitle=nips,
  year={2024}
}

@inproceedings{cheng2025simplevqa,
  title={{SimpleVQA: Multimodal Factuality Evaluation for Multimodal Large Language Models}},
  author={Cheng, Xianfu and Zhang, Wei and Zhang, Shiwei and Yang, Jian and Guan, Xiangyuan and Wu, Xianjie and Li, Xiang and Zhang, Ge and Liu, Jiaheng and Mai, Yuying and others},
  booktitle=cvpr,
  year={2025}
}

@inproceedings{shukor2025scalingnmm,
  title={{Scaling Laws for Native Multimodal Models}},
  author={Shukor, Mustafa and Fini, Enrico and da Costa, Victor Guilherme Turrisi and Cord, Matthieu and Susskind, Joshua and El-Nouby, Alaaeldin},
  booktitle=iccv,
  year={2025}
}

@article{tong2026beyondlang,
  title={{Beyond Language Modeling: An Exploration of Multimodal Pretraining}},
  author={Tong, Shengbang and Fan, David and Nguyen, John and Brown, Ellis and Zhou, Gaoyue and Qian, Shengyi and Zheng, Boyang and Vallaeys, Th{\'e}ophane and Han, Junlin and Fergus, Rob and others},
  journal={arXiv preprint arXiv:2603.03276},
  year={2026}
}

@article{kaplan2020oaiscaling,
  title={{Scaling Laws for Neural Language Models}},
  author={Kaplan, Jared and McCandlish, Sam and Henighan, Tom and Brown, Tom B and Chess, Benjamin and Child, Rewon and Gray, Scott and Radford, Alec and Wu, Jeffrey and Amodei, Dario},
  journal={arXiv preprint arXiv:2001.08361},
  year={2020}
}

@article{hoffmann2022chinchilla,
  title={{Training Compute-Optimal Large Language Models}},
  author={Hoffmann, Jordan and Borgeaud, Sebastian and Mensch, Arthur and Buchatskaya, Elena and Cai, Trevor and Rutherford, Eliza and Casas, DDL and Hendricks, Lisa Anne and Welbl, Johannes and Clark, Aidan and others},
  journal={arXiv preprint arXiv:2203.15556},
  year={2022}
}

@inproceedings{shukor2026datamixscaling,
  title={{Scaling Laws for Optimal Data Mixtures}},
  author={Shukor, Mustafa and Bethune, Louis and Busbridge, Dan and Grangier, David and Fini, Enrico and El-Nouby, Alaaeldin and Ablin, Pierre},
  booktitle=nips,
  year={2026}
}

@inproceedings{liu2023llava,
  title={{Visual Instruction Tuning}},
  author={Liu, Haotian and Li, Chunyuan and Wu, Qingyang and Lee, Yong Jae},
  booktitle=nips,
  year={2023}
}

@article{lin2026moellava,
  title={{MoE-LLaVA: Mixture of Experts for Large Vision-Language Models}},
  author={Lin, Bin and Tang, Zhenyu and Ye, Yang and Huang, Jinfa and Zhang, Junwu and Pang, Yatian and Jin, Peng and Ning, Munan and Luo, Jiebo and Yuan, Li},
  journal={IEEE Transactions on Multimedia},
  year={2026},
  publisher={IEEE}
}

@article{cui2025emu35,
  title={{Emu3.5: Native Multimodal Models are World Learners}},
  author={Cui, Yufeng and Chen, Honghao and Deng, Haoge and Huang, Xu and Li, Xinghang and Liu, Jirong and Liu, Yang and Luo, Zhuoyan and Wang, Jinsheng and Wang, Wenxuan and others},
  journal={arXiv preprint arXiv:2510.26583},
  year={2025}
}

@inproceedings{brown2020gpt3,
  title={{Language Models are Few-Shot Learners}},
  author={Brown, Tom and Mann, Benjamin and Ryder, Nick and Subbiah, Melanie and Kaplan, Jared D and Dhariwal, Prafulla and Neelakantan, Arvind and Shyam, Pranav and Sastry, Girish and Askell, Amanda and others},
  booktitle=nips,
  year={2020}
}
}

\clearpage
\appendix
\section{Implementation Details}
\label{appendix:imple}

For native multimodal pre-training, we employ the Muon optimizer with a weight decay of 0.1 and a gradient clipping threshold of 1.0. 
Additionally, we utilize a warmup-stable learning rate schedule, a global batch size of 8M tokens, and a maximum sequence length of 4096.
The training loss is computed only on text tokens, with vision tokens masked out from loss computation~\citep{kimi2026kimi25}.
We detail the specific architecture configurations and training settings in~\Cref{table:arch}.

\begin{table}[htbp]
\centering
\setlength{\tabcolsep}{8pt}
\scriptsize
\caption{\textbf{Pre-training hyperparameters.} 
We detail the hyperparameters used for pre-training different model configurations to derive scaling laws.}
\begin{tabular}{lcccccc}
\toprule
& \textbf{A71M} & \textbf{A128M} & \textbf{A340M} & \textbf{A590M} & \textbf{A874M} & \textbf{A3B} \\
\midrule
\multicolumn{7}{l}{\textbf{Architecture Configurations}} \\
\midrule
layers & 11 & 15 & 19 & 23 & 27 & 48 \\
hidden size & 640 & 768 & 1152 & 1280 & 1536 & 2048 \\
FFN hidden size & 2048 & 2048 & 3072 & 4096 & 4096 & 6912 \\
attention heads & 8 & 12 & 16 & 20 & 24 & 32 \\
query groups & \multicolumn{6}{c}{4} \\
kv channels & \multicolumn{6}{c}{128} \\
total experts & \multicolumn{6}{c}{128} \\
activated experts & \multicolumn{6}{c}{8} \\
expert FFN hidden size & 256 & 256 & 384 & 512 & 512 & 768 \\
shared-expert hidden size & 256 & 256 & 384 & 512 & 512 & 768 \\
MoE layer & \multicolumn{6}{c}{first layer dense, remaining layers MoE} \\
\midrule
\multicolumn{7}{l}{\textbf{Training Settings}} \\
\midrule
batch size (sequences) & \multicolumn{6}{c}{2048} \\
sequence length & \multicolumn{6}{c}{4096} \\
warmup steps & \multicolumn{6}{c}{2000} \\
learning rate & 2.6e-3 & 1.9e-3 & 1.1e-3 & 8.2e-4 & 6.8e-4 & 3.0e-4 \\
\bottomrule
\end{tabular}
\label{table:arch}
\end{table}

\section{Scaling-Law Fitting and Budget Compensation}
\label{appendix:scaling}

In the estimator formulas below, we suppress the objective index and mixture argument only within a fixed $(s,r)$ calculation: $C$ denotes $C_s$, $D$ denotes $D_s$, and $E,C_c,\beta$ denote $E_s(r),C_{c,s}(r),\beta_s(r)$. 
Compute is measured in FLOPs. 
The amplitude is $K_s=C_{c,s}^{\beta_s}$; it differs from the characteristic compute scale $C_{c,s}$. Allocation exponents are $a_s(r),b_s(r)$, and quadratic coefficients are $p_{j,s}$ for $j=0,1,2$. 
In the budget-compensation analysis, $C_0$ denotes the pure-text reference budget, hats denote the compute required at a specified target loss, and bars identify jointly fitted frontier parameters.
In this section, we describe the methods used to estimate the compute-optimal allocation in~\Cref{sec:scaling} and derive budget compensation from the $L_{\mathrm{text}}$ fits.

\subsection{Data Preprocessing}
\label{appendix:preprocessing}

With $T=2048\times4096\approx8\mathrm{M}$ tokens per optimizer step, the total token count at step $k$ is $D_{\mathrm{total}}=Tk$. 
Define $D_{\mathrm{text}}=D_{\mathrm{total}}/(1+r)$ and $D_{\mathrm{mm}}=rD_{\mathrm{total}}/(1+r)$, with $C_s=6ND_s$. The multimodal fits exclude $r=0$.
For each objective-mixture pair, we construct candidate budgets uniformly in $\log C_s$ between $\min_m C_{s,m}^{\min}$ and $\max_m C_{s,m}^{\max}$. At budget $C$, let
\begin{equation}
\mathcal M(C)=\{m:C_{s,m}^{\min}\leq C\leq C_{s,m}^{\max}\}.
\end{equation}
Retain only budgets with $|\mathcal M(C)|\geq3$. 

\subsection{Estimating Compute-Optimal Allocation\label{appendix:allocation_estimators}}
\minisection{IsoFLOP profiles}
We construct IsoFLOP profiles by comparing smoothed training losses across model sizes at matched compute budgets. 
At each budget $C$, we fit a quadratic in $\log N$:
\begin{equation}
L(\log N\mid C)=p_2(\log N)^2+p_1\log N+p_0.
\end{equation}
For a convex fit, the minimum gives
\begin{equation}
\log N^{\mathrm{opt}}(C)=-\frac{p_1}{2p_2},
\end{equation}
clipped to the observed range of $\log N$. 
The corresponding optimal token count is $D^{\mathrm{opt}}(C)=C/(6N^{\mathrm{opt}}(C))$. 
We then regress $\log N^{\mathrm{opt}}$ and $\log D^{\mathrm{opt}}$ on $\log C$ to estimate the allocation exponents $a$ and $b$, with $a+b=1$.

\minisection{Training-curve envelope}
We construct an equal-FLOPs envelope by selecting the lowest smoothed training loss across model sizes at each evaluated compute budget. 
Each envelope point retains its associated model size and objective-specific token count. Regressing their logarithms on $\log C$ provides a second estimate of the allocation exponents. 
Both estimators use the same runs, so their agreement provides a sensitivity check rather than independent validation. 
Their allocation exponents are compared in~\Cref{sec:scaling}.
The equal-compute envelope is
\begin{equation}
	L_s^{\mathrm{envelope}}(C;r)=\min_{m\in\mathcal M(C)}\widetilde L_{s,m}(C;r),
\end{equation}
where $\widetilde L$ is the within-model interpolated loss. 
Retain the selected model size $N$ and set $D=C/(6N)$ at each budget. 
Regressing the selected $\log N$ and $\log D$ on $\log C$ yields the envelope allocation exponents; fitting $L_s^{\mathrm{envelope}}(C;r)$ yields the distinct frontier parameters. 

\subsection{Fitting Loss-Compute Frontiers}
\label{appendix:frontier_fitting}

We fit the equal-FLOPs envelope using the power law in~\Cref{eq:power_law},
\begin{equation}
L(C)=E+(C_c/C)^{\beta},
\end{equation}
where $E$ is a fitted loss offset over the observed compute interval, $C_c$ is a characteristic compute scale, and $\beta$ is the decay exponent.
We estimate the parameters through a grid search over $\beta$.
For each candidate, we write
\begin{equation}
L(C)=E+KC^{-\beta},
\qquad K=C_c^{\beta},
\end{equation}
and solve for $(E,K)$ by ordinary least squares in loss space. 
We reject $K\leq0$ or non-positive predictions at the fitted budgets, select the candidate with the smallest mean squared log-loss residual, and recover $C_c=K^{1/\beta}$. 
This is a two-stage selection procedure rather than joint least squares in log loss; the implementation does not impose $E>0$. 
All reproduced nominal fits have positive offsets.
Each mixture is fitted independently, with its own loss offset.

\subsection{Compensation Model}
\label{appendix:compensation_models}

We jointly fit the text-loss envelopes using the model in~\Cref{eq:shared_shape_frontier}.
The offset $\overline E_{\mathrm{text}}$ and decay exponent $\overline\beta_{\mathrm{text}}$ are shared across mixtures, while $\overline C_{c,\mathrm{text}}(r)$ is estimated separately for each ratio.
The parameters minimize the unweighted squared residuals in loss space:
\begin{equation}
\min_{\theta}
\sum_i
\left[
L_{\mathrm{text},i}-\overline L_{\mathrm{text}}(C_{\mathrm{text},i};r_i)
\right]^2,
\end{equation}
where $i$ indexes the envelope observations and $\theta$ contains all fitted parameters.
For numerical optimization, we parameterize the scales by $\eta(r)=\overline\beta_{\mathrm{text}}\log[\overline C_{c,\mathrm{text}}(r)/(10^{20}\,\mathrm{FLOPs})]$, with bounds $\overline E_{\mathrm{text}}\in[0,2.2]$, $\overline\beta_{\mathrm{text}}\in[0.01,0.6]$, and $\eta(r)\in[-10,5]$.
No ordering constraint is imposed across mixtures.
The fitted compute scales determine the compensation factors through~\Cref{eq:added_token_coverage}; compensation is not imposed during fitting.  
Each factor describes an evaluated mixture, without assuming a continuous dependence on $r$.

\section{Experimental Results}
\label{appendix:exp}

This section reports per-benchmark results supporting~\Cref{sec:exp} in three groups.
First, we present the training dynamics of the A3B model, evaluating text and multimodal performance at successive pre-training token budgets across multimodal data ratios $r \in \{0, 0.1, 0.2, 0.3\}$ (\Cref{table:pt_text_mm0,table:pt_text_mm25,table:pt_text_mm50,table:pt_text_mm75,table:pt_mm25,table:pt_mm50,table:pt_mm75}).
Second, we report the final text performance across the entire model family, scaling from A71M to A3B, for each data ratio (\Cref{table:text_mm0,table:text_mm25,table:text_mm50,table:text_mm75}). 
Third, we evaluate multimodal in-context learning, providing per-benchmark accuracies for the 0-, 1-, and 3-shot settings across all model capacities (\Cref{table:fewshot_0shot,table:fewshot_1shot,table:fewshot_3shot}). 
All aggregate scores represent unweighted means across the constituent benchmarks. 
The few-shot tables average all 23 multimodal tasks, whereas the curves in~\Cref{sec:exp} average 21 tasks, excluding ScienceQA and LogicVista because their CoT evaluations are unstable without in-context examples.

\clearpage

\begin{table}[!tb]
\centering
\setlength\tabcolsep{7.775pt}
\scriptsize
\caption{\textbf{Performance on text benchmarks at different pre-training token budgets for A3B models. The models are trained with 250B text tokens.}}
\begin{tabular}{l|l|cccccccccc}
\toprule
& Total tokens $D_{\mathrm{total}}$ & 25B & 50B & 75B & 100B & 125B & 150B & 175B & 200B & 225B & 250B \\
\midrule
\multirow{4}{*}{\makecell[l]{Aggregate}} 
& MMLU-Redux (Acc.) & 27.07 & 41.26 & 46.77 & 49.86 & 52.39 & 53.02 & 54.35 & 55.65 & 56.42 & 57.79 \\
& MMLU-Pro (Acc.) & 10.80 & 14.25 & 16.68 & 17.64 & 20.10 & 21.28 & 21.86 & 24.09 & 24.26 & 24.88 \\
& AGIEval$_{\mathrm{en}}$ (Acc.) & 19.53 & 26.20 & 28.55 & 30.70 & 34.21 & 34.36 & 36.26 & 35.74 & 35.13 & 36.96 \\
& SuperGPQA (Acc.) & 10.00 & 13.29 & 14.71 & 17.00 & 16.70 & 18.83 & 20.16 & 19.81 & 20.33 & 19.95 \\
\midrule
\multirow{2}{*}{\makecell[l]{Coding}} 
& HumanEval+ (Pass@1) & 9.15 & 12.80 & 13.41 & 16.46 & 14.63 & 18.90 & 16.46 & 14.63 & 18.90 & 20.73 \\
& MBPP+ (Pass@1) & 17.72 & 34.92 & 34.66 & 40.48 & 47.09 & 48.68 & 47.35 & 50.26 & 51.59 & 48.68 \\
\midrule
\multirow{2}{*}{\makecell[l]{Mathematics}} 
& GSM8K (Pass@1) & 6.52 & 16.22 & 27.60 & 35.78 & 35.25 & 39.80 & 43.90 & 46.02 & 49.73 & 48.75 \\
& MATH (Pass@1) & 5.40 & 10.30 & 14.90 & 17.40 & 19.10 & 20.10 & 22.10 & 21.55 & 25.00 & 26.70 \\
\midrule
\multirow{2}{*}{\makecell[l]{Logical \\ Reasoning}} 
& BBH (EM) & 29.84 & 30.08 & 34.46 & 35.56 & 38.66 & 40.26 & 41.06 & 43.39 & 39.76 & 44.97 \\
& SpatialEval (Acc.) & 27.70 & 33.04 & 31.74 & 34.59 & 29.94 & 34.80 & 38.19 & 37.93 & 40.28 & 35.95 \\
\midrule
\multirow{2}{*}{\makecell[l]{Knowledge}} 
& NaturalQuestions (EM) & 10.42 & 15.24 & 18.70 & 20.72 & 20.19 & 20.11 & 22.94 & 22.63 & 22.91 & 22.74 \\
& TriviaQA (EM) & 21.81 & 33.33 & 40.42 & 46.94 & 47.08 & 51.25 & 52.36 & 54.58 & 54.58 & 56.11 \\
\midrule
\multirow{4}{*}{\makecell[l]{Commonsense \\ Reasoning}} 
& HellaSwag (Acc.) & 52.73 & 61.00 & 65.10 & 66.80 & 68.83 & 69.07 & 70.40 & 70.87 & 71.67 & 71.57 \\
& SIQA (Acc.) & 33.52 & 47.34 & 52.56 & 56.29 & 56.35 & 58.60 & 58.24 & 59.72 & 61.26 & 60.24 \\
& PIQA (Acc.) & 71.27 & 73.83 & 74.97 & 76.39 & 77.04 & 77.86 & 77.20 & 78.67 & 77.97 & 78.35 \\
& WinoGrande (Acc.) & 52.88 & 55.56 & 56.27 & 56.83 & 57.06 & 60.06 & 60.77 & 60.38 & 59.83 & 61.96 \\
\midrule
\rowcolor{blue!5} \multicolumn{2}{c|}{Average} & 25.40 & 32.42 & 35.72 & 38.71 & 39.66 & 41.69 & 42.73 & 43.49 & 44.35 & 44.77 \\
\bottomrule
\end{tabular}
\label{table:pt_text_mm0}
\end{table}

\begin{table}[!tb]
\centering
\setlength\tabcolsep{6.6pt}
\scriptsize
\caption{\textbf{Performance on text benchmarks at different pre-training token budgets for A3B models. The models are trained with 250B text tokens and 25B multimodal tokens.}}
\begin{tabular}{l|l|ccccccccccc}
\toprule
& Total tokens $D_{\mathrm{total}}$ & 25B & 50B & 75B & 100B & 125B & 150B & 175B & 200B & 225B & 250B & 275B \\
\midrule
\multirow{4}{*}{\makecell[l]{Aggregate}} 
& MMLU-Redux (Acc.) & 28.65 & 37.00 & 44.88 & 49.72 & 51.23 & 52.51 & 53.54 & 54.14 & 54.74 & 55.54 & 56.84 \\
& MMLU-Pro (Acc.) & 11.14 & 13.41 & 16.67 & 17.92 & 20.34 & 22.11 & 21.71 & 23.07 & 24.71 & 25.42 & 25.29 \\
& AGIEval$_{\mathrm{en}}$ (Acc.) & 21.15 & 23.49 & 26.53 & 30.15 & 33.32 & 33.24 & 33.11 & 33.78 & 35.16 & 35.60 & 37.38 \\
& SuperGPQA (Acc.) & 10.03 & 12.77 & 15.42 & 15.84 & 18.07 & 17.89 & 18.56 & 19.21 & 19.87 & 20.77 & 20.31 \\
\midrule
\multirow{2}{*}{\makecell[l]{Coding}} 
& HumanEval+ (Pass@1) & 6.71 & 6.10 & 11.59 & 14.02 & 12.80 & 20.73 & 18.90 & 19.51 & 18.29 & 17.07 & 19.51 \\
& MBPP+ (Pass@1) & 16.14 & 28.57 & 34.13 & 40.48 & 43.12 & 46.03 & 48.15 & 50.53 & 51.59 & 52.65 & 53.97 \\
\midrule
\multirow{2}{*}{\makecell[l]{Mathematics}} 
& GSM8K (Pass@1) & 5.00 & 14.86 & 25.47 & 33.36 & 38.51 & 38.44 & 42.91 & 48.98 & 48.22 & 48.67 & 50.49 \\
& MATH (Pass@1) & 7.20 & 11.75 & 14.60 & 17.20 & 19.60 & 21.15 & 23.20 & 24.85 & 25.40 & 25.85 & 25.60 \\
\midrule
\multirow{2}{*}{\makecell[l]{Logical \\ Reasoning}} 
& BBH (EM) & 26.54 & 33.29 & 34.45 & 36.66 & 38.85 & 38.71 & 41.83 & 42.59 & 44.25 & 43.08 & 43.60 \\
& SpatialEval (Acc.) & 26.68 & 32.37 & 32.31 & 31.39 & 38.87 & 39.63 & 38.52 & 39.34 & 38.65 & 39.78 & 39.28 \\
\midrule
\multirow{2}{*}{\makecell[l]{Knowledge}} 
& NaturalQuestions (EM) & 10.69 & 15.32 & 18.84 & 18.78 & 19.45 & 21.22 & 23.41 & 24.76 & 24.27 & 24.71 & 23.49 \\
& TriviaQA (EM) & 21.11 & 33.33 & 38.06 & 44.44 & 47.64 & 49.44 & 51.11 & 55.42 & 55.97 & 54.44 & 57.92 \\
\midrule
\multirow{4}{*}{\makecell[l]{Commonsense \\ Reasoning}} 
& HellaSwag (Acc.) & 51.47 & 61.50 & 64.43 & 66.87 & 68.03 & 69.13 & 69.97 & 70.53 & 71.37 & 70.83 & 71.30 \\
& SIQA (Acc.) & 35.11 & 45.55 & 50.51 & 54.96 & 52.92 & 58.96 & 55.94 & 55.48 & 59.16 & 56.86 & 62.54 \\
& PIQA (Acc.) & 71.33 & 75.03 & 75.30 & 76.33 & 76.88 & 77.53 & 75.95 & 77.97 & 77.75 & 78.29 & 77.80 \\
& WinoGrande (Acc.) & 52.01 & 56.83 & 57.46 & 58.72 & 58.64 & 57.85 & 59.51 & 61.01 & 62.83 & 62.12 & 62.67 \\
\midrule
\rowcolor{blue!5} \multicolumn{2}{c|}{Average} & 25.06 & 31.32 & 35.04 & 37.93 & 39.89 & 41.54 & 42.27 & 43.82 & 44.51 & 44.48 & 45.50 \\
\bottomrule
\end{tabular}
\label{table:pt_text_mm25}
\end{table}

\begin{table}[!tb]
\centering
\setlength\tabcolsep{5.625pt}
\scriptsize
\caption{\textbf{Performance on text benchmarks at different pre-training token budgets for A3B models. The models are trained with 250B text tokens and 50B multimodal tokens.}}
\begin{tabular}{l|l|cccccccccccc}
\toprule
& Total tokens $D_{\mathrm{total}}$ & 25B & 50B & 75B & 100B & 125B & 150B & 175B & 200B & 225B & 250B & 275B & 300B \\
\midrule
\multirow{4}{*}{\makecell[l]{Aggregate}} 
& MMLU-Redux (Acc.) & 26.19 & 40.39 & 46.39 & 48.00 & 49.81 & 52.39 & 52.46 & 54.39 & 55.60 & 56.09 & 55.89 & 56.68 \\
& MMLU-Pro (Acc.) & 11.28 & 12.58 & 17.01 & 18.24 & 18.23 & 19.75 & 21.88 & 21.05 & 21.17 & 19.85 & 24.18 & 23.93 \\
& AGIEval$_{\mathrm{en}}$ (Acc.) & 21.48 & 23.11 & 28.85 & 29.50 & 31.59 & 32.48 & 34.79 & 33.80 & 33.56 & 34.25 & 36.55 & 36.32 \\
& SuperGPQA (Acc.) & 10.17 & 12.64 & 14.06 & 16.02 & 15.74 & 17.50 & 18.31 & 18.43 & 19.22 & 20.55 & 20.98 & 20.52 \\
\midrule
\multirow{2}{*}{\makecell[l]{Coding}} 
& HumanEval+ (Pass@1) & 7.93 & 11.59 & 14.63 & 17.68 & 18.29 & 18.90 & 21.95 & 20.12 & 20.73 & 19.51 & 18.90 & 20.12 \\
& MBPP+ (Pass@1) & 14.55 & 26.19 & 33.33 & 42.33 & 42.86 & 47.09 & 47.35 & 46.56 & 48.94 & 48.41 & 50.79 & 51.85 \\
\midrule
\multirow{2}{*}{\makecell[l]{Mathematics}} 
& GSM8K (Pass@1) & 5.61 & 15.24 & 23.50 & 30.33 & 37.07 & 36.01 & 44.66 & 49.28 & 50.27 & 52.92 & 50.57 & 53.90 \\
& MATH (Pass@1) & 5.25 & 10.00 & 14.25 & 16.20 & 17.25 & 19.85 & 21.30 & 22.50 & 22.95 & 25.30 & 24.15 & 25.35 \\
\midrule
\multirow{2}{*}{\makecell[l]{Logical \\ Reasoning}}
& BBH (EM) & 27.59 & 27.13 & 31.29 & 37.16 & 38.27 & 40.99 & 40.33 & 40.12 & 42.04 & 43.09 & 44.10 & 45.44 \\
& SpatialEval (Acc.) & 30.41 & 33.65 & 37.61 & 36.87 & 35.04 & 32.48 & 33.13 & 34.72 & 37.37 & 34.89 & 35.91 & 38.04 \\
\midrule
\multirow{2}{*}{\makecell[l]{Knowledge}} 
& NaturalQuestions (EM) & 10.89 & 14.21 & 17.31 & 20.83 & 20.33 & 21.16 & 22.11 & 20.33 & 22.33 & 21.88 & 21.16 & 21.44 \\
& TriviaQA (EM) & 19.72 & 27.64 & 34.31 & 42.22 & 47.64 & 48.89 & 49.31 & 54.72 & 51.67 & 56.53 & 56.67 & 57.50 \\
\midrule
\multirow{4}{*}{\makecell[l]{Commonsense \\ Reasoning}} 
& HellaSwag (Acc.) & 51.67 & 59.77 & 63.67 & 66.90 & 67.60 & 69.23 & 68.60 & 70.30 & 70.77 & 71.10 & 72.00 & 71.60 \\
& SIQA (Acc.) & 33.88 & 42.94 & 45.29 & 53.22 & 54.55 & 56.24 & 56.40 & 56.96 & 59.21 & 58.39 & 60.59 & 60.29 \\
& PIQA (Acc.) & 70.95 & 74.32 & 75.46 & 76.99 & 77.04 & 77.91 & 78.13 & 79.22 & 78.40 & 78.62 & 78.18 & 78.73 \\
& WinoGrande (Acc.) & 52.49 & 55.41 & 55.72 & 59.43 & 59.83 & 60.38 & 61.64 & 61.64 & 61.56 & 64.01 & 62.67 & 61.64 \\
\midrule
\rowcolor{blue!5} \multicolumn{2}{c|}{Average} & 25.00 & 30.43 & 34.54 & 38.24 & 39.45 & 40.70 & 42.02 & 42.76 & 43.49 & 44.09 & 44.58 & 45.21 \\
\bottomrule
\end{tabular}
\label{table:pt_text_mm50}
\end{table}

\begin{table}[!tb]
\centering
\setlength\tabcolsep{4.675pt}
\scriptsize
\caption{\textbf{Performance on text benchmarks at different pre-training token budgets for A3B models. The models are trained with 250B text tokens and 75B multimodal tokens.}}
\begin{tabular}{l|l|ccccccccccccc}
\toprule
& Total tokens $D_{\mathrm{total}}$ & 25B & 50B & 75B & 100B & 125B & 150B & 175B & 200B & 225B & 250B & 275B & 300B & 325B \\
\midrule
\multirow{4}{*}{\makecell[l]{Aggregate}} 
& MMLU-Redux (Acc.) & 27.05 & 38.81 & 44.74 & 48.35 & 51.19 & 51.37 & 54.49 & 54.21 & 55.05 & 56.00 & 57.18 & 57.09 & 57.98 \\
& MMLU-Pro (Acc.) & 10.85 & 12.99 & 17.73 & 19.19 & 19.86 & 20.21 & 22.03 & 23.84 & 24.61 & 24.92 & 25.33 & 27.53 & 27.68 \\
& AGIEval$_{\mathrm{en}}$ (Acc.) & 20.49 & 22.71 & 27.88 & 29.27 & 32.13 & 32.66 & 36.03 & 36.58 & 35.02 & 35.99 & 37.11 & 36.92 & 37.73 \\
& SuperGPQA (Acc.) & 10.78 & 13.48 & 14.92 & 16.88 & 18.11 & 18.27 & 18.69 & 19.63 & 20.17 & 20.63 & 21.08 & 21.53 & 22.00 \\
\midrule
\multirow{2}{*}{\makecell[l]{Coding}} 
& HumanEval+ (Pass@1) & 6.10 & 11.59 & 15.24 & 14.63 & 15.24 & 15.24 & 13.42 & 18.90 &  19.51 & 15.85 & 17.68 & 21.34 & 21.34 \\
& MBPP+ (Pass@1) & 18.52 & 26.98 & 31.48 & 37.83 & 41.27 & 47.09 & 46.03 & 48.94 & 49.47 & 48.94 & 49.74 & 53.44 & 54.23 \\
\midrule
\multirow{2}{*}{\makecell[l]{Mathematics}} 
& GSM8K (Pass@1) & 4.85 & 17.21 & 22.21 & 30.63 & 37.00 & 36.01 & 42.15 & 43.82 & 48.07 & 50.19 & 49.58 & 51.40 & 52.39 \\
& MATH (Pass@1) & 5.95 & 10.60 & 14.05 & 16.05 & 19.00 & 19.70 & 20.65 & 21.80 & 22.60 & 23.90 & 24.65 & 25.00 & 26.30 \\
\midrule
\multirow{2}{*}{\makecell[l]{Logical \\ Reasoning}} 
& BBH (EM) & 29.21 & 29.44 & 33.50 & 33.93 & 34.28 & 36.56 & 40.07 & 39.49 & 39.08 & 42.94 & 43.82 & 43.68 & 40.84 \\
& SpatialEval (Acc.) & 29.85 & 32.59 & 36.78 & 34.52 & 36.98 & 37.24 & 39.02 & 35.85 & 39.04 & 41.91 & 40.95 & 44.25 & 41.43 \\
\midrule
\multirow{2}{*}{\makecell[l]{Knowledge}} 
& NaturalQuestions (EM) & 10.08 & 14.99 & 16.32 & 18.17 & 19.31 & 21.77 & 24.13 & 22.05 & 19.64 & 22.02 & 23.35 & 21.55 & 23.63 \\
& TriviaQA (EM) & 20.56 & 32.08 & 37.08 & 42.78 & 43.47 & 46.81 & 50.42 & 50.97 & 52.92 & 54.58 & 57.22 & 58.19 & 57.92 \\
\midrule
\multirow{4}{*}{\makecell[l]{Commonsense \\ Reasoning}} 
& HellaSwag (Acc.) & 51.23 & 59.40 & 63.20 & 66.47 & 66.97 & 67.87 & 68.27 & 69.37 & 69.53 & 70.60 & 71.23 & 71.37 & 71.57 \\
& SIQA (Acc.) & 34.85 & 45.14 & 52.71 & 54.15 & 56.24 & 55.48 & 59.72 & 59.67 & 60.90 & 62.28 & 62.08 & 61.87 & 62.95 \\
& PIQA (Acc.) & 71.16 & 74.59 & 75.63 & 76.12 & 77.31 & 76.88 & 77.42 & 77.31 & 78.29 & 77.86 & 79.05 & 78.45 & 77.80 \\
& WinoGrande (Acc.) & 50.83 & 55.41 & 56.43 & 59.12 & 59.51 & 58.25 & 59.98 & 61.64 & 59.27 & 62.12 & 61.25 & 61.17 & 60.62 \\
\midrule
\rowcolor{blue!5} \multicolumn{2}{c|}{Average} & 25.15 & 31.13 & 34.99 & 37.38 & 39.24 & 40.09 & 42.03 & 42.75 & 43.32 & 44.42 & 45.08 & 45.92 & 46.03 \\
\bottomrule
\end{tabular}
\label{table:pt_text_mm75}
\end{table}

\begin{table}[!tb]
\centering
\setlength\tabcolsep{6.6pt}
\scriptsize
\caption{\textbf{Performance on multimodal benchmarks at different pre-training token budgets for A3B models. The models are trained with 250B text tokens and 25B multimodal tokens.}}
\begin{tabular}{l|l|ccccccccccc}
\toprule
& Total tokens $D_{\mathrm{total}}$ & 25B & 50B & 75B & 100B & 125B & 150B & 175B & 200B & 225B & 250B & 275B \\
\midrule
\multirow{5}{*}{\makecell[l]{Aggregate}}
& MMStar (Acc.) & 26.65 & 25.62 & 25.21 & 27.07 & 28.37 & 25.21 & 26.72 & 28.65 & 26.31 & 29.75 & 28.72 \\
& MMMU (Acc.) & 25.59 & 30.81 & 35.28 & 33.42 & 32.92 & 33.79 & 33.66 & 36.02 & 34.04 & 36.89 & 35.53 \\
& MMMU-Pro (Acc.) & 13.29 & 14.97 & 15.30 & 14.97 & 15.44 & 17.92 & 17.85 & 18.59 & 18.52 & 19.93 & 19.19 \\
& MME (Acc.) & 50.44 & 50.84 & 50.71 & 52.56 & 53.01 & 55.04 & 55.79 & 53.09 & 56.50 & 56.01 & 54.51 \\
& MMBench$_{\mathrm{en}}$ (Acc.) & 29.71 & 32.84 & 35.37 & 36.67 & 38.08 & 38.64 & 39.41 & 40.01 & 41.44 & 41.46 & 42.86 \\
\midrule
\multirow{2}{*}{\makecell[l]{VQA}}
& VQAv2 (EM) & 39.84 & 42.74 & 42.14 & 43.86 & 44.46 & 45.48 & 45.72 & 47.22 & 45.60 & 46.74 & 49.20 \\
& TextVQA (EM) & 10.66 & 10.58 & 9.34 & 10.98 & 9.58 & 11.52 & 11.92 & 13.02 & 14.10 & 13.08 & 13.86 \\
\midrule
\multirow{3}{*}{\makecell[l]{STEM}}
& MathVista (Pass@1) & 34.00 & 33.60 & 37.20 & 38.00 & 39.80 & 42.40 & 39.20 & 40.80 & 42.20 & 41.60 & 41.00 \\
& MathVerse (Pass@1) & 29.53 & 29.06 & 30.84 & 31.67 & 33.10 & 33.87 & 29.89 & 33.87 & 33.69 & 33.99 & 33.39 \\
& ScienceQA (EM) & 34.01 & 43.58 & 45.46 & 49.68 & 53.10 & 52.75 & 53.30 & 52.45 & 54.64 & 49.23 & 56.72 \\
\midrule
\multirow{4}{*}{\makecell[l]{Doc \\ Understanding}}
& HallusionBench (Acc.) & 49.23 & 46.58 & 43.38 & 42.96 & 42.40 & 44.49 & 41.98 & 42.40 & 44.63 & 43.38 & 46.03 \\
& LogicVista (EM) & 15.85 & 20.73 & 23.17 & 23.48 & 23.17 & 27.13 & 21.95 & 23.48 & 26.52 & 26.52 & 28.66 \\
& AI2D (Acc.) & 24.18 & 37.11 & 42.80 & 45.88 & 49.21 & 48.85 & 50.88 & 53.63 & 51.28 & 52.95 & 52.49 \\
& ChartQA (Acc.) & 2.67 & 4.58 & 5.11 & 4.98 & 5.92 & 4.29 & 4.58 & 5.19 & 4.66 & 5.47 & 6.16 \\
\midrule
\multirow{2}{*}{\makecell[l]{Vision \\ Knowledge}} 
& MMBench$_{\mathrm{cc}}$ (EM) & 26.97 & 26.72 & 27.32 & 30.56 & 29.34 & 31.62 & 31.87 & 30.3 & 31.26 & 30.96 & 31.67 \\
& SimpleVQA (EM) & 5.87 & 6.77 & 7.12 & 8.48 & 8.23 & 9.38 & 8.58 & 9.78 & 10.49 & 10.29 & 8.98 \\
\midrule
\multirow{2}{*}{\makecell[l]{Counting}} 
& CountBench (EM) & 12.39 & 13.91 & 11.74 & 15.43 & 16.52 & 13.91 & 12.39 & 15.22 & 16.09 & 17.83 & 16.74 \\
& CountQA (EM) & 2.89 & 7.84 & 6.40 & 3.10 & 3.10 & 4.44 & 5.16 & 3.92 & 5.37 & 5.88 & 3.82 \\
\midrule
\multirow{5}{*}{\makecell[l]{Spatial \\ Reasoning}} 
& RealWorldQA (Acc.) & 35.58 & 38.60 & 39.30 & 39.30 & 43.49 & 37.91 & 41.63 & 43.72 & 41.63 & 37.67 & 41.40 \\
& CV-Bench (Acc.) & 40.79 & 44.51 & 45.93 & 44.17 & 45.36 & 47.62 & 46.51 & 46.51 & 46.51 & 46.55 & 45.32 \\
& OmniSpatial (Acc.) & 19.45 & 33.03 & 26.28 & 27.75 & 35.39 & 34.42 & 31.49 & 36.37 & 34.26 & 34.66 & 40.52 \\
& SEAM (Acc.) & 22.45 & 25.51 & 27.55 & 31.22 & 30.37 & 29.45 & 31.73 & 31.52 & 30.94 & 30.54 & 31.15 \\
& SpatialEval (Acc.) & 27.44 & 32.65 & 30.55 & 28.16 & 37.63 & 38.61 & 34.13 & 32.80 & 35.74 & 37.39 & 33.85 \\
\midrule
\rowcolor{orange!5} \multicolumn{2}{c|}{Average} & 25.19 & 28.40 & 28.85 & 29.75 & 31.22 & 31.68 & 31.15 & 32.11 & 32.45 & 32.56 & 33.12 \\
\bottomrule
\end{tabular}
\label{table:pt_mm25}
\end{table}

\begin{table}[!tb]
\centering
\setlength\tabcolsep{5.57pt}
\scriptsize
\caption{\textbf{Performance on multimodal benchmarks at different pre-training token budgets for A3B models. The models are trained with 250B text tokens and 50B multimodal tokens.}}
\begin{tabular}{l|l|cccccccccccc}
\toprule
& Total tokens $D_{\mathrm{total}}$ & 25B & 50B & 75B & 100B & 125B & 150B & 175B & 200B & 225B & 250B & 275B & 300B \\
\midrule
\multirow{5}{*}{\makecell[l]{Aggregate}} 
& MMStar (Acc.) & 28.37 & 24.38 & 28.93 & 27.82 & 26.72 & 25.83 & 29.96 & 27.82 & 28.79 & 30.23 & 29.96 & 28.79 \\
& MMMU (Acc.) & 29.19 & 30.19 & 31.93 & 34.66 & 37.64 & 37.27 & 37.39 & 36.89 & 34.53 & 37.02 & 36.52 & 37.64 \\
& MMMU-Pro (Acc.) & 13.22 & 14.09 & 14.63 & 15.57 & 16.17 & 17.05 & 18.19 & 16.11 & 18.12 & 18.52 & 18.66 & 18.59 \\
& MME (Acc.) & 51.59 & 49.29 & 52.03 & 52.92 & 51.24 & 51.90 & 57.74 & 51.28 & 60.21 & 56.72 & 53.80 & 57.65 \\
& MMBench$_{\mathrm{en}}$ (Acc.) & 28.40 & 33.54 & 35.88 & 38.36 & 40.18 & 42.30 & 42.37 & 44.73 & 47.12 & 48.10 & 48.00 & 48.49 \\
\midrule
\multirow{2}{*}{\makecell[l]{VQA}} 
& VQAv2 (EM) & 41.98 & 43.16 & 47.68 & 46.76 & 50.04 & 48.88 & 50.18 & 47.88 & 50.20 & 52.10 & 52.28 & 50.18 \\
& TextVQA (EM) & 11.54 & 11.52 & 12.66 & 13.66 & 13.34 & 14.06 & 14.54 & 16.80 & 21.50 & 22.62 & 25.10 & 27.98 \\
\midrule
\multirow{3}{*}{\makecell[l]{STEM}} 
& MathVista (Pass@1) & 35.80 & 38.40 & 38.40 & 40.00 & 38.80 & 42.00 & 40.80 & 40.60 & 41.00 & 40.20 & 40.20 & 39.40 \\
& MathVerse (Pass@1) & 29.00 & 27.45 & 31.49 & 33.63 & 34.46 & 33.87 & 32.62 & 33.93 & 30.36 & 34.11 & 34.82 & 34.40 \\
& ScienceQA (EM) & 31.18 & 41.99 & 47.25 & 49.33 & 52.55 & 47.60 & 49.33 & 51.26 & 53.79 & 55.23 & 55.28 & 57.06 \\
\midrule
\multirow{4}{*}{\makecell[l]{Doc \\ Understanding}} 
& HallusionBench (Acc.) & 50.07 & 42.12 & 41.56 & 41.84 & 42.82 & 48.40 & 45.19 & 44.91 & 43.10 & 41.56 & 43.65 & 47.84 \\
& LogicVista (EM) & 18.60 & 21.34 & 24.09 & 23.48 & 23.48 & 25.00 & 19.82 & 21.95 & 25.30 & 23.17 & 25.30 & 27.13 \\
& AI2D (Acc.) & 23.79 & 34.88 & 44.18 & 46.79 & 47.19 & 49.54 & 51.96 & 51.73 & 51.21 & 51.44 & 51.15 & 52.78 \\
& ChartQA (Acc.) & 4.54 & 3.89 & 4.50 & 4.25 & 4.29 & 5.71 & 4.54 & 5.47 & 3.97 & 5.75 & 5.19 & 4.94 \\
\midrule
\multirow{2}{*}{\makecell[l]{Vision \\ Knowledge}} 
& MMBench$_{\mathrm{cc}}$ (EM) & 26.62 & 26.41 & 26.52 & 26.92 & 29.49 & 30.45 & 33.13 & 31.72 & 32.42 & 33.13 & 32.47 & 33.13 \\
& SimpleVQA (EM) & 6.72 & 6.87 & 7.98 & 9.33 & 8.63 & 9.68 & 8.53 & 8.43 & 10.24 & 11.69 & 12.39 & 10.69 \\
\midrule
\multirow{2}{*}{\makecell[l]{Counting}} 
& CountBench (EM) & 11.96 & 13.48 & 12.61 & 14.78 & 15.65 & 16.52 & 17.39 & 20.65 & 21.74 & 22.39 & 28.70 & 28.26 \\
& CountQA (EM) & 7.84 & 6.40 & 3.72 & 3.41 & 4.64 & 6.19 & 5.16 & 7.53 & 7.02 & 5.47 & 4.33 & 5.16 \\
\midrule
\multirow{5}{*}{\makecell[l]{Spatial \\ Reasoning}} 
& RealWorldQA (Acc.) & 33.95 & 36.28 & 42.09 & 43.26 & 41.86 & 47.21 & 42.09 & 45.35 & 42.09 & 39.07 & 41.63 & 43.49 \\
& CV-Bench (Acc.) & 41.63 & 39.10 & 41.52 & 43.90 & 44.32 & 46.12 & 43.21 & 42.48 & 43.94 & 46.28 & 45.28 & 47.39 \\
& OmniSpatial (Acc.) & 21.40 & 31.49 & 32.79 & 34.42 & 32.47 & 35.23 & 33.93 & 38.97 & 38.49 & 35.96 & 36.37 & 42.72 \\
& SEAM (Acc.) & 23.06 & 24.63 & 25.03 & 25.88 & 25.85 & 25.61 & 28.53 & 26.49 & 31.01 & 30.20 & 29.55 & 27.79 \\
& SpatialEval (Acc.) & 28.33 & 29.76 & 36.04 & 36.74 & 33.11 & 33.50 & 32.78 & 33.04 & 35.72 & 35.37 & 32.91 & 36.24 \\
\midrule
\rowcolor{orange!5} \multicolumn{2}{c|}{Average} & 26.03 & 27.42 & 29.72 & 30.77 & 31.08 & 32.17 & 32.15 & 32.44 & 33.56 & 33.75 & 34.07 & 35.12 \\
\bottomrule
\end{tabular}
\label{table:pt_mm50}
\end{table}

\begin{table}[!tb]
\centering
\setlength\tabcolsep{4.7pt}
\scriptsize
\caption{\textbf{Performance on multimodal benchmarks at different pre-training token budgets for A3B models. The models are trained with 250B text tokens and 75B multimodal tokens.}}
\begin{tabular}{l|l|ccccccccccccc}
\toprule
& Total tokens $D_{\mathrm{total}}$ & 25B & 50B & 75B & 100B & 125B & 150B & 175B & 200B & 225B & 250B & 275B & 300B & 325B \\
\midrule
\multirow{5}{*}{\makecell[l]{Aggregate}} 
& MMStar (Acc.) & 28.99 & 25.96 & 28.72 & 27.89 & 30.65 & 29.13 & 29.48 & 30.85 & 28.51 & 32.16 & 33.33 & 31.13 & 30.51 \\
& MMMU (Acc.) & 29.07 & 29.94 & 30.19 & 33.66 & 34.66 & 35.90 & 33.79 & 33.91 & 35.90 & 37.27 & 36.02 & 34.91 & 37.64 \\
& MMMU-Pro (Acc.) & 12.01 & 13.89 & 14.83 & 16.11 & 16.91 & 18.05 & 17.25 & 18.72 & 18.66 & 18.46 & 18.86 & 19.87 & 21.54 \\
& MME (Acc.) & 51.41 & 50.04 & 51.41 & 51.72 & 50.49 & 53.89 & 57.34 & 57.65 & 61.80 & 61.98 & 62.16 & 58.36 & 59.86 \\
& MMBench$_{\mathrm{en}}$ (Acc.) & 28.47 & 33.12 & 36.93 & 38.92 & 40.43 & 41.77 & 45.55 & 48.80 & 49.05 & 52.25 & 51.97 & 52.28 & 54.73 \\
\midrule
\multirow{2}{*}{\makecell[l]{VQA}} 
& VQAv2 (EM) & 40.70 & 46.74 & 46.32 & 47.08 & 49.68 & 49.72 & 49.60 & 49.92 & 49.66 & 51.16 & 49.42 & 53.70 & 62.78 \\
& TextVQA (EM) & 11.88 & 13.78 & 12.94 & 12.28 & 13.86 & 16.56 & 21.46 & 25.00 & 30.26 & 30.76 & 33.76 & 34.52 & 39.70 \\
\midrule
\multirow{3}{*}{\makecell[l]{STEM}} 
& MathVista (Pass@1) & 34.00 & 33.40 & 37.20 & 37.00 & 39.60 & 41.00 & 42.20 & 41.00 & 41.80 & 44.40 & 41.20 & 41.20 & 43.00 \\
& MathVerse (Pass@1) & 23.35 & 29.95 & 27.93 & 30.07 & 28.58 & 33.10 & 34.52 & 32.20 & 31.97 & 36.07 & 33.87 & 33.33 & 36.13 \\
& ScienceQA (EM) & 34.16 & 39.56 & 46.31 & 48.04 & 49.23 & 54.88 & 51.36 & 52.75 & 54.39 & 53.79 & 55.03 & 55.28 & 57.16 \\
\midrule
\multirow{4}{*}{\makecell[l]{Doc \\ Understanding}} 
& HallusionBench (Acc.) & 51.60 & 41.70 & 42.68 & 41.56 & 41.70 & 45.61 & 42.26 & 41.14 & 41.56 & 42.26 & 42.82 & 48.81 & 43.93 \\
& LogicVista (EM) & 18.29 & 19.82 & 20.73 & 21.65 & 21.65 & 22.56 & 21.65 & 24.70 & 19.82 & 19.82 & 25.00 & 20.43 & 25.61 \\
& AI2D (Acc.) & 24.74 & 34.82 & 43.78 & 46.56 & 48.69 & 48.82 & 51.05 & 52.62 & 52.68 & 53.89 & 54.22 & 51.77 & 54.45 \\
& ChartQA (Acc.) & 4.42 & 3.48 & 3.73 & 4.90 & 4.54 & 4.21 & 4.38 & 4.17 & 5.43 & 6.32 & 6.69 & 7.33 & 8.06 \\
\midrule
\multirow{2}{*}{\makecell[l]{Vision \\ Knowledge}} 
& MMBench$_{\mathrm{cc}}$ (EM) & 24.80 & 26.16 & 29.09 & 30.00 & 32.42 & 31.92 & 32.17 & 33.94 & 35.00 & 34.95 & 33.28 & 36.77 & 36.87 \\
& SimpleVQA (EM) & 7.12 & 8.73 & 8.98 & 8.93 & 10.44 & 9.48 & 10.34 & 11.14 & 11.74 & 11.39 & 11.29 & 13.15 & 13.95 \\
\midrule
\multirow{2}{*}{\makecell[l]{Counting}}
& CountBench (EM) & 10.43 & 10.87 & 15.65 & 11.30 & 15.22 & 25.22 & 23.04 & 24.78 & 28.91 & 27.83 & 30.22 & 34.78 & 34.13 \\
& CountQA (EM) & 3.72 & 1.55 & 4.95 & 4.13 & 5.16 & 2.68 & 7.33 & 5.88 & 6.09 & 5.47 & 6.19 & 7.43 & 7.53 \\
\midrule
\multirow{5}{*}{\makecell[l]{Spatial \\ Reasoning}} 
& RealWorldQA (Acc.) & 39.53 & 37.67 & 40.00 & 40.23 & 43.49 & 43.26 & 40.23 & 36.98 & 41.16 & 42.56 & 39.77 & 42.79 & 44.88 \\
& CV-Bench (Acc.) & 41.90 & 45.78 & 44.90 & 44.01 & 43.98 & 41.10 & 42.90 & 42.56 & 44.97 & 45.32 & 45.43 & 46.05 & 46.39 \\
& OmniSpatial (Acc.) & 21.07 & 24.57 & 35.31 & 36.94 & 34.50 & 36.05 & 33.93 & 32.79 & 37.43 & 40.93 & 37.27 & 35.39 & 37.02 \\
& SEAM (Acc.) & 21.91 & 23.91 & 25.20 & 28.57 & 25.44 & 28.19 & 28.02 & 28.06 & 28.67 & 30.98 & 28.46 & 28.63 & 32.10 \\
& SpatialEval (Acc.) & 32.46 & 31.96 & 32.74 & 35.69 & 34.63 & 31.70 & 33.98 & 30.35 & 35.76 & 37.58 & 36.65 & 37.80 & 34.33 \\
\midrule
\rowcolor{orange!5} \multicolumn{2}{c|}{Average} & 25.91 & 27.28 & 29.59 & 30.31 & 31.13 & 32.38 & 32.78 & 33.04 & 34.40 & 35.55 & 35.34 & 35.90 & 37.49 \\
\bottomrule
\end{tabular}
\label{table:pt_mm75}
\end{table}

\clearpage

\begin{table}[!tb]
\centering
\scriptsize
\setlength\tabcolsep{6pt}
\caption{\textbf{Performance on text benchmarks. Model sizes ($N$) range from $71$M to $3$B activated parameters, with all models trained on 250B text tokens.}}
\label{table:text_mm0}
\begin{tabular}{l|l|cccccc}
\toprule
& Model Sizes ($N$) & A71M & A128M & A340M & A590M & A874M & A3B \\
\midrule
\multirow{4}{*}{\makecell[l]{Aggregate}} 
& MMLU-Redux (Acc.) & 26.39 & 31.75 & 40.79 & 44.74 & 49.72 & 57.79 \\
& MMLU-Pro (Acc.) & 11.47 & 12.39 & 13.95 & 14.11 & 21.46 & 24.88 \\
& AGIEval$_{\mathrm{en}}$ (Acc.) & 20.72 & 18.98 & 24.61 & 28.81 & 31.56 & 36.96 \\
& SuperGPQA (Acc.) & 9.99 & 10.36 & 12.88 & 13.04 & 17.12 & 19.95 \\
\midrule
\multirow{2}{*}{\makecell[l]{Coding}} 
& HumanEval+ (Pass@1) & 4.88 & 7.93 & 15.24 & 15.24 & 18.90 & 20.73 \\
& MBPP+ (Pass@1) & 11.64 & 16.67 & 28.31 & 35.19 & 39.68 & 48.68 \\
\midrule
\multirow{2}{*}{\makecell[l]{Mathematics}} 
& GSM8K (Pass@1) & 1.90 & 5.91 & 15.31 & 22.97 & 25.02 & 48.75 \\
& MATH (Pass@1) & 3.30 & 5.15 & 10.10 & 13.40 & 16.05 & 26.70 \\
\midrule
\multirow{2}{*}{\makecell[l]{Logical \\ Reasoning}} 
& BBH (EM) & 26.78 & 27.71 & 30.77 & 32.75 & 35.15 & 44.97 \\
& SpatialEval (Acc.) & 28.24 & 27.89 & 26.40 & 29.87 & 32.65 & 35.95 \\
\midrule
\multirow{2}{*}{\makecell[l]{Knowledge}} 
& NaturalQuestions (EM) & 7.76 & 10.66 & 12.96 & 15.62 & 17.89 & 22.74 \\
& TriviaQA (EM) & 12.64 & 19.58 & 26.11 & 35.56 & 40.14 & 56.11 \\
\midrule
\multirow{4}{*}{\makecell[l]{Commonsense \\ Reasoning}} 
& HellaSwag (Acc.) & 42.57 & 47.73 & 56.97 & 62.57 & 64.80 & 71.57 \\
& SIQA (Acc.) & 34.29 & 36.34 & 44.22 & 44.78 & 53.58 & 60.24 \\
& PIQA (Acc.) & 68.88 & 70.29 & 73.67 & 74.10 & 76.17 & 78.35 \\
& WinoGrande (Acc.) & 52.49 & 50.12 & 54.54 & 55.33 & 56.83 & 61.96 \\
\midrule
\rowcolor{blue!5} \multicolumn{2}{c|}{Average} & 22.75 & 24.97 & 30.43 & 33.63 & 37.30 & 44.77 \\
\bottomrule
\end{tabular}
\end{table}

\begin{table}[!tb]
\centering
\scriptsize
\setlength\tabcolsep{6pt}
\caption{\textbf{Performance on text benchmarks. Model sizes ($N$) range from $71$M to $3$B activated parameters, with all models trained on 250B text tokens and 25B multimodal tokens.}}
\label{table:text_mm25}
\begin{tabular}{l|l|cccccc}
\toprule
& Model Sizes ($N$) & A71M & A128M & A340M & A590M & A874M & A3B \\
\midrule
\multirow{4}{*}{\makecell[l]{Aggregate}} 
& MMLU-Redux (Acc.) & 28.11 & 32.09 & 42.14 & 46.95 & 49.60 & 56.84 \\
& MMLU-Pro (Acc.) & 11.51 & 11.33 & 14.27 & 15.66 & 17.32 & 25.29 \\
& AGIEval$_{\mathrm{en}}$ (Acc.) & 21.81 & 21.36 & 24.33 & 27.63 & 29.07 & 37.38 \\
& SuperGPQA (Acc.) & 8.79 & 9.34 & 13.10 & 13.35 & 14.62 & 20.31 \\
\midrule
\multirow{2}{*}{\makecell[l]{Coding}} 
& HumanEval+ (Pass@1) & 6.71 & 9.76 & 13.41 & 16.46 & 15.85 & 19.51 \\
& MBPP+ (Pass@1) & 7.14 & 12.96 & 29.10 & 33.86 & 38.89 & 53.97 \\
\midrule
\multirow{2}{*}{\makecell[l]{Mathematics}} 
& GSM8K (Pass@1) & 1.82 & 5.53 & 12.51 & 20.02 & 28.66 & 50.49 \\
& MATH (Pass@1) & 3.45 & 5.00 & 8.50 & 14.35 & 16.25 & 25.60 \\
\midrule
\multirow{2}{*}{\makecell[l]{Logical \\ Reasoning}} 
& BBH (EM) & 25.34 & 27.14 & 30.38 & 32.85 & 34.34 & 43.60 \\
& SpatialEval (Acc.) & 28.35 & 26.80 & 35.35 & 32.93 & 34.96 & 39.28 \\
\midrule
\multirow{2}{*}{\makecell[l]{Knowledge}} 
& NaturalQuestions (EM) & 7.98 & 9.42 & 13.80 & 15.24 & 17.37 & 23.49 \\
& TriviaQA (EM) & 13.89 & 16.11 & 24.31 & 34.72 & 41.53 & 57.92 \\
\midrule
\multirow{4}{*}{\makecell[l]{Commonsense \\ Reasoning}} 
& HellaSwag (Acc.) & 42.47 & 46.67 & 57.47 & 62.90 & 65.33 & 71.30 \\
& SIQA (Acc.) & 34.29 & 34.80 & 42.43 & 45.80 & 52.71 & 62.54 \\
& PIQA (Acc.) & 69.31 & 69.31 & 73.45 & 73.94 & 76.28 & 77.80 \\
& WinoGrande (Acc.) & 50.51 & 51.54 & 55.41 & 56.91 & 57.85 & 62.67 \\
\midrule
\rowcolor{blue!5} \multicolumn{2}{c|}{Average} & 22.59 & 24.32 & 30.62 & 33.97 & 36.91 & 45.50 \\
\bottomrule
\end{tabular}
\end{table}

\begin{table}[!tb]
\centering
\scriptsize
\setlength\tabcolsep{6pt}
\caption{\textbf{Performance on text benchmarks. Model sizes ($N$) range from $71$M to $3$B activated parameters, with all models trained on 250B text tokens and 50B multimodal tokens.}}
\label{table:text_mm50}
\begin{tabular}{l|l|cccccc}
\toprule
& Model Sizes ($N$) & A71M & A128M & A340M & A590M & A874M & A3B \\
\midrule
\multirow{4}{*}{\makecell[l]{Aggregate}} 
& MMLU-Redux (Acc.) & 26.28 & 32.72 & 41.09 & 47.02 & 48.61 & 56.68 \\
& MMLU-Pro (Acc.) & 11.24 & 11.73 & 13.33 & 15.61 & 21.19 & 23.93 \\
& AGIEval$_{\mathrm{en}}$ (Acc.) & 21.11 & 22.78 & 25.50 & 27.27 & 32.58 & 36.32 \\
& SuperGPQA (Acc.) & 9.47 & 9.82 & 13.09 & 13.73 & 17.17 & 20.52 \\
\midrule
\multirow{2}{*}{\makecell[l]{Coding}} 
& HumanEval+ (Pass@1) & 7.32 & 7.93 & 12.80 & 15.85 & 17.07 & 20.12 \\
& MBPP+ (Pass@1) & 8.73 & 12.17 & 26.72 & 34.13 & 39.15 & 51.85 \\
\midrule
\multirow{2}{*}{\makecell[l]{Mathematics}} 
& GSM8K (Pass@1) & 2.58 & 4.25 & 14.10 & 24.56 & 27.60 & 53.90 \\
& MATH (Pass@1) & 2.90 & 5.90 & 10.15 & 13.00 & 15.50 & 25.35 \\
\midrule
\multirow{2}{*}{\makecell[l]{Logical \\ Reasoning}} 
& BBH (EM) & 27.00 & 28.71 & 30.27 & 32.04 & 35.77 & 45.44 \\
& SpatialEval (Acc.) & 27.76 & 25.05 & 30.18 & 36.06 & 37.95 & 38.04 \\
\midrule
\multirow{2}{*}{\makecell[l]{Knowledge}} 
& NaturalQuestions (EM) & 7.42 & 9.78 & 13.88 & 16.18 & 17.48 & 21.44 \\
& TriviaQA (EM) & 15.56 & 19.17 & 28.89 & 34.44 & 40.42 & 57.50 \\
\midrule
\multirow{4}{*}{\makecell[l]{Commonsense \\ Reasoning}} 
& HellaSwag (Acc.) & 42.70 & 47.90 & 57.67 & 61.97 & 64.90 & 71.60 \\
& SIQA (Acc.) & 36.03 & 35.72 & 43.24 & 49.49 & 53.89 & 60.29 \\
& PIQA (Acc.) & 66.76 & 69.97 & 73.61 & 74.48 & 76.61 & 78.73 \\
& WinoGrande (Acc.) & 51.14 & 51.93 & 54.54 & 55.41 & 57.54 & 61.64 \\
\midrule
\rowcolor{blue!5} \multicolumn{2}{c|}{Average} & 22.75 & 24.72 & 30.57 & 34.45 & 37.71 & 45.21 \\
\bottomrule
\end{tabular}
\end{table}

\begin{table}[!tb]
\centering
\scriptsize
\setlength\tabcolsep{6pt}
\caption{\textbf{Performance on text benchmarks. Model sizes ($N$) range from $71$M to $3$B activated parameters, with all models trained on 250B text tokens and 75B multimodal tokens.}}
\label{table:text_mm75}
\begin{tabular}{l|l|cccccc}
\toprule
& Model Sizes ($N$) & A71M & A128M & A340M & A590M & A874M & A3B \\
\midrule
\multirow{4}{*}{\makecell[l]{Aggregate}} 
& MMLU-Redux (Acc.) & 25.40 & 32.11 & 38.91 & 44.95 & 49.04 & 57.98 \\
& MMLU-Pro (Acc.) & 10.68 & 11.63 & 13.87 & 14.97 & 17.28 & 27.68 \\
& AGIEval$_{\mathrm{en}}$ (Acc.) & 20.64 & 20.45 & 25.54 & 26.29 & 29.67 & 37.73 \\
& SuperGPQA (Acc.) & 8.00 & 10.31 & 11.81 & 14.16 & 14.90 & 22.00 \\
\midrule
\multirow{2}{*}{\makecell[l]{Coding}} 
& HumanEval+ (Pass@1) & 4.88 & 7.93 & 12.20 & 16.46 & 15.85 & 21.34 \\
& MBPP+ (Pass@1) & 7.94 & 13.23 & 26.46 & 30.95 & 37.57 & 54.23 \\
\midrule
\multirow{2}{*}{\makecell[l]{Mathematics}} 
& GSM8K (Pass@1) & 2.27 & 5.08 & 14.25 & 23.28 & 28.05 & 52.39 \\
& MATH (Pass@1) & 3.70 & 4.95 & 9.40 & 11.95 & 15.60 & 26.30 \\
\midrule
\multirow{2}{*}{\makecell[l]{Logical \\ Reasoning}} 
& BBH (EM) & 26.87 & 28.06 & 30.31 & 33.03 & 35.93 & 40.84 \\
& SpatialEval (Acc.) & 27.94 & 30.17 & 32.39 & 38.93 & 33.07 & 41.43 \\
\midrule
\multirow{2}{*}{\makecell[l]{Knowledge}} 
& NaturalQuestions (EM) & 7.15 & 10.03 & 13.85 & 14.65 & 18.86 & 23.63 \\
& TriviaQA (EM) & 17.08 & 17.92 & 25.97 & 36.94 & 39.86 & 57.92 \\
\midrule
\multirow{4}{*}{\makecell[l]{Commonsense \\ Reasoning}} 
& HellaSwag (Acc.) & 42.13 & 48.13 & 57.57 & 62.10 & 65.60 & 71.57 \\
& SIQA (Acc.) & 32.45 & 33.88 & 43.04 & 49.90 & 52.05 & 62.95 \\
& PIQA (Acc.) & 68.50 & 69.64 & 73.45 & 74.05 & 75.68 & 77.80 \\
& WinoGrande (Acc.) & 51.54 & 51.70 & 53.83 & 56.35 & 58.41 & 60.62 \\
\midrule
\rowcolor{blue!5} \multicolumn{2}{c|}{Average} & 22.32 & 24.70 & 30.18 & 34.31 & 36.71 & 46.03 \\
\bottomrule
\end{tabular}
\end{table}


\begin{table}[!tb]
\centering
\scriptsize
\setlength\tabcolsep{6pt}
\caption{\textbf{Performance on multimodal benchmarks in a 0-shot setting. Model sizes ($N$) range from $71$M to $3$B activated parameters, with all models trained on 250B text tokens and 75B multimodal tokens.}}
\label{table:fewshot_0shot}
\begin{tabular}{l|l|cccccc}
\toprule
& Model Sizes ($N$) & A71M & A128M & A340M & A590M & A874M & A3B \\
\midrule
\multirow{5}{*}{\makecell[l]{Aggregate}}
& MMStar (Acc.) & 27.82 & 26.86 & 24.59 & 28.58 & 28.86 & 30.17 \\
& MMMU (Acc.) & 28.70 & 24.97 & 28.07 & 31.80 & 30.31 & 36.65 \\
& MMMU-Pro (Acc.) & 12.42 & 14.30 & 13.62 & 15.37 & 15.17 & 16.98 \\
& MME (Acc.) & 49.73 & 50.18 & 50.71 & 51.90 & 53.58 & 50.35 \\
& MMBench$_\mathrm{en}$ (Acc.) & 27.12 & 30.79 & 34.95 & 43.42 & 47.26 & 54.10 \\
\midrule
\multirow{2}{*}{\makecell[l]{VQA}} 
& VQAv2 (EM) & 45.44 & 46.78 & 51.70 & 48.52 & 52.24 & 58.32 \\
& TextVQA (EM) & 14.92 & 14.54 & 16.02 & 24.90 & 29.28 & 41.24 \\
\midrule
\multirow{3}{*}{\makecell[l]{STEM}}  
& MathVista (Pass@1) & 33.40 & 32.00 & 31.00 & 37.20 & 37.80 & 39.00 \\
& MathVerse (Pass@1) & 27.63 & 29.71 & 21.87 & 30.36 & 28.34 & 34.17 \\
& ScienceQA-CoT (EM) & 0.00 & 0.84 & 12.99 & 18.94 & 7.88 & 18.54 \\
\midrule
\multirow{4}{*}{\makecell[l]{Doc \\ Understanding}} 
& HallusionBench (Acc.) & 52.30 & 44.63 & 49.65 & 41.98 & 41.14 & 47.84 \\
& LogicVista-CoT (EM) & 0.91 & 6.40 & 5.18 & 18.60 & 6.71 & 5.18 \\
& AI2D (Acc.) & 25.82 & 26.77 & 31.15 & 39.04 & 40.18 & 47.02 \\
& ChartQA (Acc.) & 4.46 & 4.90 & 5.39 & 6.36 & 5.96 & 8.87 \\
\midrule
\multirow{2}{*}{\makecell[l]{Vision \\ Knowledge}} 
& MMBench$_\mathrm{cc}$ (EM) & 25.76 & 25.96 & 28.33 & 28.74 & 33.33 & 33.03 \\
& SimpleVQA (EM) & 3.51 & 4.92 & 5.92 & 8.08 & 8.83 & 9.08 \\
\midrule
\multirow{2}{*}{\makecell[l]{Counting}} 
& CountBench (EM) & 8.70 & 13.91 & 15.22 & 20.43 & 20.43 & 34.13 \\
& CountQA (EM) & 5.88 & 3.41 & 3.92 & 5.47 & 2.06 & 5.37 \\
\midrule
\multirow{5}{*}{\makecell[l]{Spatial \\ Reasoning}}  
& RealWorldQA (Acc.) & 34.65 & 41.16 & 39.53 & 41.16 & 38.84 & 41.86 \\
& CV-Bench (Acc.) & 41.67 & 42.36 & 42.36 & 41.71 & 44.24 & 41.48 \\
& OmniSpatial (Acc.) & 21.89 & 20.67 & 22.86 & 23.76 & 23.92 & 22.78 \\
& SEAM (Acc.) & 24.69 & 25.92 & 26.26 & 25.37 & 26.87 & 27.38 \\
& SpatialEval (Acc.) & 26.85 & 24.29 & 27.20 & 26.85 & 22.59 & 27.26 \\
\midrule
\rowcolor{orange!5} \multicolumn{2}{c|}{Average} & 23.66 & 24.19 & 25.59 & 28.63 & 28.08 & 31.77 \\
\bottomrule
\end{tabular}
\end{table}

\begin{table}[!tb]
\centering
\scriptsize
\setlength\tabcolsep{6pt}
\caption{\textbf{Performance on multimodal benchmarks in a 1-shot setting. Model sizes ($N$) range from $71$M to $3$B activated parameters, with all models trained on 250B text tokens and 75B multimodal tokens.}}
\label{table:fewshot_1shot}
\begin{tabular}{l|l|cccccc}
\toprule
& Model Sizes ($N$) & A71M & A128M & A340M & A590M & A874M & A3B \\
\midrule
\multirow{5}{*}{\makecell[l]{Aggregate}}  
& MMStar (Acc.) & 28.79 & 29.48 & 27.34 & 27.69 & 27.34 & 31.20 \\
& MMMU (Acc.) & 27.58 & 30.31 & 33.17 & 31.30 & 37.14 & 38.76 \\
& MMMU-Pro (Acc.) & 12.28 & 14.23 & 14.23 & 14.43 & 14.83 & 20.94 \\
& MME (Acc.) & 50.57 & 49.69 & 50.13 & 51.06 & 50.62 & 58.53 \\
& MMBench$_\mathrm{en}$ (Acc.) & 27.80 & 30.88 & 33.17 & 38.68 & 41.91 & 52.00 \\
\midrule
\multirow{2}{*}{\makecell[l]{VQA}} 
& VQAv2 (EM) & 42.06 & 45.92 & 46.74 & 47.34 & 48.64 & 62.78 \\
& TextVQA (EM) & 13.18 & 14.66 & 13.18 & 21.60 & 22.16 & 39.70 \\
\midrule
\multirow{3}{*}{\makecell[l]{STEM}} 
& MathVista (Pass@1) & 34.20 & 34.60 & 39.80 & 38.80 & 39.60 & 40.60 \\
& MathVerse (Pass@1) & 31.02 & 30.30 & 28.64 & 30.48 & 32.03 & 33.21 \\
& ScienceQA-CoT (EM) & 34.01 & 27.62 & 43.38 & 43.78 & 48.29 & 54.78 \\
\midrule
\multirow{4}{*}{\makecell[l]{Doc \\ Understanding}} 
& HallusionBench (Acc.) & 44.77 & 41.28 & 43.38 & 41.28 & 41.70 & 49.23 \\
& LogicVista-CoT (EM) & 20.73 & 23.48 & 18.90 & 21.65 & 23.78 & 25.61 \\
& AI2D (Acc.) & 24.51 & 29.65 & 37.04 & 41.03 & 45.22 & 53.53 \\
& ChartQA (Acc.) & 2.35 & 3.44 & 3.53 & 4.74 & 4.29 & 8.35 \\
\midrule
\multirow{2}{*}{\makecell[l]{Vision \\ Knowledge}} 
& MMBench$_\mathrm{cc}$ (EM) & 26.77 & 25.66 & 27.78 & 29.80 & 30.76 & 33.54 \\
& SimpleVQA (EM) & 4.26 & 3.56 & 5.02 & 6.07 & 6.17 & 8.78 \\
\midrule
\multirow{2}{*}{\makecell[l]{Counting}} 
& CountBench (EM) & 10.00 & 12.83 & 12.83 & 17.83 & 18.48 & 23.04 \\
& CountQA (EM) & 7.64 & 5.26 & 4.44 & 8.67 & 5.78 & 7.53 \\
\midrule
\multirow{5}{*}{\makecell[l]{Spatial \\ Reasoning}}
& RealWorldQA (Acc.) & 32.79 & 37.44 & 35.81 & 43.26 & 42.79 & 39.30 \\
& CV-Bench (Acc.) & 40.94 & 38.33 & 46.16 & 41.52 & 45.28 & 46.51 \\
& OmniSpatial (Acc.) & 23.43 & 23.76 & 31.57 & 34.74 & 37.10 & 36.13 \\
& SEAM (Acc.) & 22.32 & 24.80 & 23.88 & 23.81 & 26.09 & 30.67 \\
& SpatialEval (Acc.) & 27.66 & 27.61 & 26.55 & 36.28 & 32.80 & 34.33 \\
\midrule
\rowcolor{orange!5} \multicolumn{2}{c|}{Average} & 25.64 & 26.30 & 28.12 & 30.25 & 31.43 & 36.05 \\
\bottomrule
\end{tabular}
\end{table}

\begin{table}[!tb]
\centering
\scriptsize
\setlength\tabcolsep{6pt}
\caption{\textbf{Performance on multimodal benchmarks in a 3-shot setting. Model sizes ($N$) range from $71$M to $3$B activated parameters, with all models trained on 250B text tokens and 75B multimodal tokens.}}

\label{table:fewshot_3shot}
\begin{tabular}{l|l|cccccc}
\toprule
& Model Sizes ($N$) & A71M & A128M & A340M & A590M & A874M & A3B \\
\midrule
\multirow{5}{*}{\makecell[l]{Aggregate}} 
& MMStar (Acc.) & 29.48 & 25.00 & 27.13 & 27.20 & 28.51 & 30.51 \\
& MMMU (Acc.) & 26.21 & 30.06 & 30.43 & 31.43 & 33.66 & 37.64 \\
& MMMU-Pro (Acc.) & 12.48 & 14.30 & 13.36 & 15.17 & 15.84 & 21.54 \\
& MME (Acc.) & 50.53 & 50.66 & 50.49 & 52.30 & 52.79 & 59.86 \\
& MMBench$_\mathrm{en}$ (Acc.) & 27.26 & 30.83 & 33.47 & 40.64 & 43.87 & 54.73 \\
\midrule
\multirow{2}{*}{\makecell[l]{VQA}} 
& VQAv2 (EM) & 42.58 & 45.42 & 47.94 & 48.50 & 51.90 & 57.28 \\
& TextVQA (EM) & 16.64 & 17.22 & 16.96 & 24.72 & 29.42 & 38.42 \\
\midrule
\multirow{3}{*}{\makecell[l]{STEM}} 
& MathVista (Pass@1) & 34.00 & 34.60 & 39.00 & 38.40 & 39.00 & 43.00 \\
& MathVerse (Pass@1) & 29.59 & 29.89 & 29.59 & 31.19 & 29.53 & 36.13 \\
& ScienceQA-CoT (EM) & 30.19 & 42.34 & 44.72 & 48.93 & 48.64 & 57.16 \\
\midrule
\multirow{4}{*}{\makecell[l]{Doc \\ Understanding}} 
& HallusionBench (Acc.) & 42.26 & 42.68 & 41.56 & 41.28 & 41.42 & 43.93 \\
& LogicVista-CoT (EM) & 23.17 & 24.39 & 13.11 & 21.34 & 24.09 & 25.61 \\
& AI2D (Acc.) & 24.71 & 30.96 & 38.68 & 42.15 & 45.98 & 54.45 \\
& ChartQA (Acc.) & 2.76 & 4.25 & 4.62 & 6.12 & 5.67 & 8.06 \\
\midrule
\multirow{2}{*}{\makecell[l]{Vision \\ Knowledge}} 
& MMBench$_\mathrm{cc}$ (EM) & 26.57 & 26.01 & 28.08 & 30.91 & 32.63 & 36.87 \\
& SimpleVQA (EM) & 6.22 & 6.62 & 6.97 & 9.13 & 9.03 & 13.95 \\
\midrule
\multirow{2}{*}{Counting} 
& CountBench (EM) & 11.74 & 12.39 & 13.48 & 19.35 & 18.48 & 23.70 \\
& CountQA (EM) & 7.95 & 6.81 & 7.12 & 6.71 & 7.84 & 6.81 \\
\midrule
\multirow{5}{*}{\makecell[l]{Spatial \\ Reasoning}} 
& RealWorldQA (Acc.) & 36.98 & 41.86 & 39.77 & 41.63 & 44.88 & 44.88 \\
& CV-Bench (Acc.) & 41.75 & 39.98 & 46.51 & 42.40 & 43.28 & 46.39 \\
& OmniSpatial (Acc.) & 22.05 & 25.87 & 31.49 & 33.60 & 36.70 & 37.02 \\
& SEAM (Acc.) & 23.61 & 24.35 & 23.78 & 25.58 & 29.11 & 32.10 \\
& SpatialEval (Acc.) & 27.48 & 29.48 & 27.40 & 33.15 & 29.55 & 30.89 \\
\midrule
\rowcolor{orange!5} \multicolumn{2}{c|}{Average} & 25.92 & 27.65 & 28.51 & 30.95 & 32.25 & 36.56 \\
\bottomrule
\end{tabular}
\end{table}

\end{document}